\documentclass[conference]{IEEEtran}
\IEEEoverridecommandlockouts
\usepackage{cite}
\usepackage{amsmath,amssymb,amsfonts}
\newtheorem{definition}{Definition}
\usepackage{algorithmic}
\usepackage{graphicx}
\usepackage{textcomp}
\usepackage{xcolor}
\usepackage{caption}
\usepackage{subcaption}
\usepackage{booktabs}
\usepackage{multirow}
\usepackage{array}
\usepackage{bbm}
\def\BibTeX{{\rm B\kern-.05em{\sc i\kern-.025em b}\kern-.08em
    T\kern-.1667em\lower.7ex\hbox{E}\kern-.125emX}}
\begin{document}

\title{Unraveling the `Anomaly' in Time Series Anomaly Detection: A Self-supervised Tri-domain Solution\\
\thanks{}
}

\author{\IEEEauthorblockN{Yuting Sun\IEEEauthorrefmark{2},
Guansong Pang\IEEEauthorrefmark{3}, Guanhua Ye\IEEEauthorrefmark{4}, 
Tong Chen\IEEEauthorrefmark{2},
Xia Hu\IEEEauthorrefmark{5} and 
Hongzhi Yin\IEEEauthorrefmark{2}\IEEEauthorrefmark{1}\thanks{* Corresponding author.}}

\IEEEauthorblockA{\IEEEauthorrefmark{2}School of Electrical Engineering and Computer Science,
the University of Queensland\\
yuting.sun@uqconnect.edu.au ~~ tong.chen@uq.edu.au ~~ h.yin1@uq.edu.au\\
\IEEEauthorrefmark{3}School of Computing and Information Systems, Singapore Management University\\
gspang@smu.edu.sg\\
\IEEEauthorrefmark{4}Research and Development Department, Deep Neural Computing Company Limited\\
rex.ye@dncc.tech\\
\IEEEauthorrefmark{5}Department of Computer Science, Rice University\\
xia.hu@rice.edu}}
\maketitle

% \author{
% \IEEEauthorblockN{1\textsuperscript{st} Yuting Sun}
% \IEEEauthorblockA{\textit{the University of Queensland}\\
% Brisbane, Australia \\
% yuting.sun@uqconnect.edu.au}
% \and
% \IEEEauthorblockN{2\textsuperscript{nd} Guansong Pang}
% \IEEEauthorblockA{\textit{Singapore Management University}\\
% Singapore \\
% gspang@smu.edu.sg}
% \and
% \IEEEauthorblockN{3\textsuperscript{rd} Guanhua Ye}
% \IEEEauthorblockA{\textit{Deep Neural Computing Company Limited}\\
% China \\
% rex.ye@dncc.tech}
% \and
% \IEEEauthorblockN{4\textsuperscript{th} Tong Chen}
% \IEEEauthorblockA{\textit{the University of Queensland}\\
% Brisbane, Australia \\
% tong.chen@uq.edu.au}
% \and
% \IEEEauthorblockN{5\textsuperscript{th} Xia Hu}
% \IEEEauthorblockA{\textit{Rice University}\\
% Houston, United States \\
% xia.hu@rice.edu}
% \and
% \IEEEauthorblockN{6\textsuperscript{th} Hongzhi Yin}
% \IEEEauthorblockA{
% \textit{the University of Queensland}\\
% Brisbane, Australia \\
% h.yin1@uq.edu.au}
% }
% \maketitle
\begin{abstract}
The ongoing challenges in time series anomaly detection (TSAD), including the scarcity of anomaly labels and the variability in anomaly lengths and shapes, have led to the need for a more robust and efficient solution. As limited anomaly labels hinder traditional supervised models in anomaly detection, various state-of-the-art (SOTA) deep learning techniques (e.g., self-supervised learning) are introduced to tackle this issue. However, they encounter difficulties handling variations in anomaly lengths and shapes, limiting their adaptability to diverse anomalies. Additionally, many benchmark datasets suffer from the problem of having explicit anomalies that even random functions can detect. This problem is exacerbated by an ill-posed evaluation metric, known as point adjustment (PA), which results in inflated model performance. In this context, we propose a novel self-supervised learning based \textbf{\underline{Tri}}-domain \textbf{\underline{A}}nomaly \textbf{\underline{D}}etector (\textbf{TriAD}), which addresses these challenges by modeling features across three aspects - temporal, frequency, and residual domains - without relying on anomaly labels. Unlike traditional contrastive learning methods, TriAD employs both inter-domain and intra-domain contrastive loss to learn common attributes among normal data and differentiate them from anomalies. Additionally, our approach can detect anomalies of varying lengths by integrating with a discord discovery algorithm. It is worth noting that this study is the first to reevaluate the deep learning potential in TSAD, utilizing both rigorously designed datasets (i.e., UCR Archive) and evaluation metrics (i.e., PA\%K and affiliation). Experimental results demonstrate that TriAD achieves an impressive three-fold increase in PA\%K based F1 scores over SOTA deep learning models. Moreover, in comparison to SOTA discord discovery algorithms, TriAD improves anomaly detection accuracy by 50\% while cutting the inference time down to just one-tenth. Illuminating the significance of rigorous datasets and evaluation metrics, this paper offers a new direction for addressing the multifaceted challenges of time series anomaly detection. The source code is
publicly available at https://github.com/pseudo-Skye/TriAD.
\end{abstract}


\section{Introduction}
Identifying anomalies within time series data plays a crucial role in a broad range of fields, including health surveillance systems, financial services, and manufacturing processes. These anomalies may represent critical incidents like sleep apnea syndrome\cite{ye2021fenet}, fraudulent transactions\cite{jiang2023anomaly}, and supply chain hiccups\cite{ABOUTORAB2022110}, each exhibiting distinct characteristics. Despite numerous efforts to address this task from different perspectives, there are still multiple challenges that hinder the progress of robust anomaly detection methods. 

\begin{figure}[t!]
    \centering
     \begin{subfigure}[b]{\columnwidth}
         \centering
         \includegraphics[width=\linewidth]{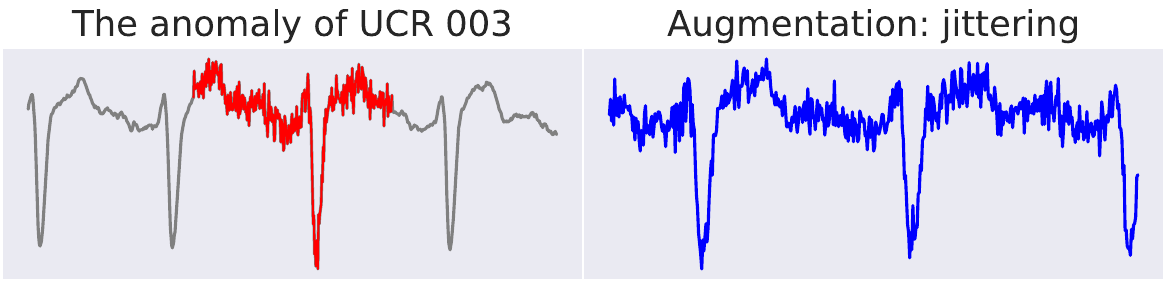}
     \end{subfigure}
     \par\bigskip
     \begin{subfigure}[b]{\columnwidth}
         \centering
         \includegraphics[width=\linewidth]{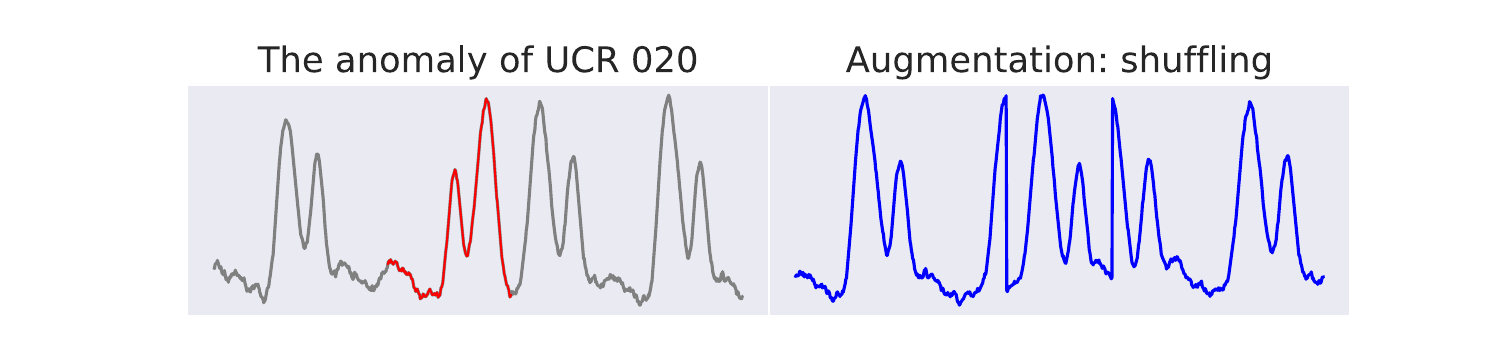}
     \end{subfigure}
\caption{Traditional data augmentation used in TSAD can make the augmented data look like anomalies.}
        \label{fig:ucr_anom}
\end{figure}

One of the primary challenges in time series anomaly detection (TSAD) is the scarcity or unavailability of the labels of anomaly cases. To address this limitation, one leading approach is self-supervised learning \cite{jiao2022timeautoad, martini2021deep, wwwDeldari}, which effectively leverages the unlabeled data and makes the most of the available annotations. However, many methods grounded in contrastive learning \cite{ijcai2021p324, 9914650, ZHOU2022266} rely on data augmentation techniques such as jittering, shuffling, or scaling to create positive pairs of augmented samples. This approach is borrowed from computer vision and unsuited for anomaly detection in time series data. This is because, as shown in Fig.~\ref{fig:ucr_anom}, the altered characteristics introduced during augmentation can be interpreted as anomalies in real-world datasets, leading to potential inaccuracies in the detection process.

% \begin{figure}[t!]
%   \centering
%   \captionsetup{justification=centering}
%   \includegraphics[width= 0.9\linewidth]{ucr_anomaly.pdf}
%   \caption{The anomalies in UCR test set.}
%   \label{fig:ucr_anom}
% \end{figure}

\begin{figure}[t!]
  \centering
  \captionsetup{justification=centering}
  \includegraphics[width= \linewidth]{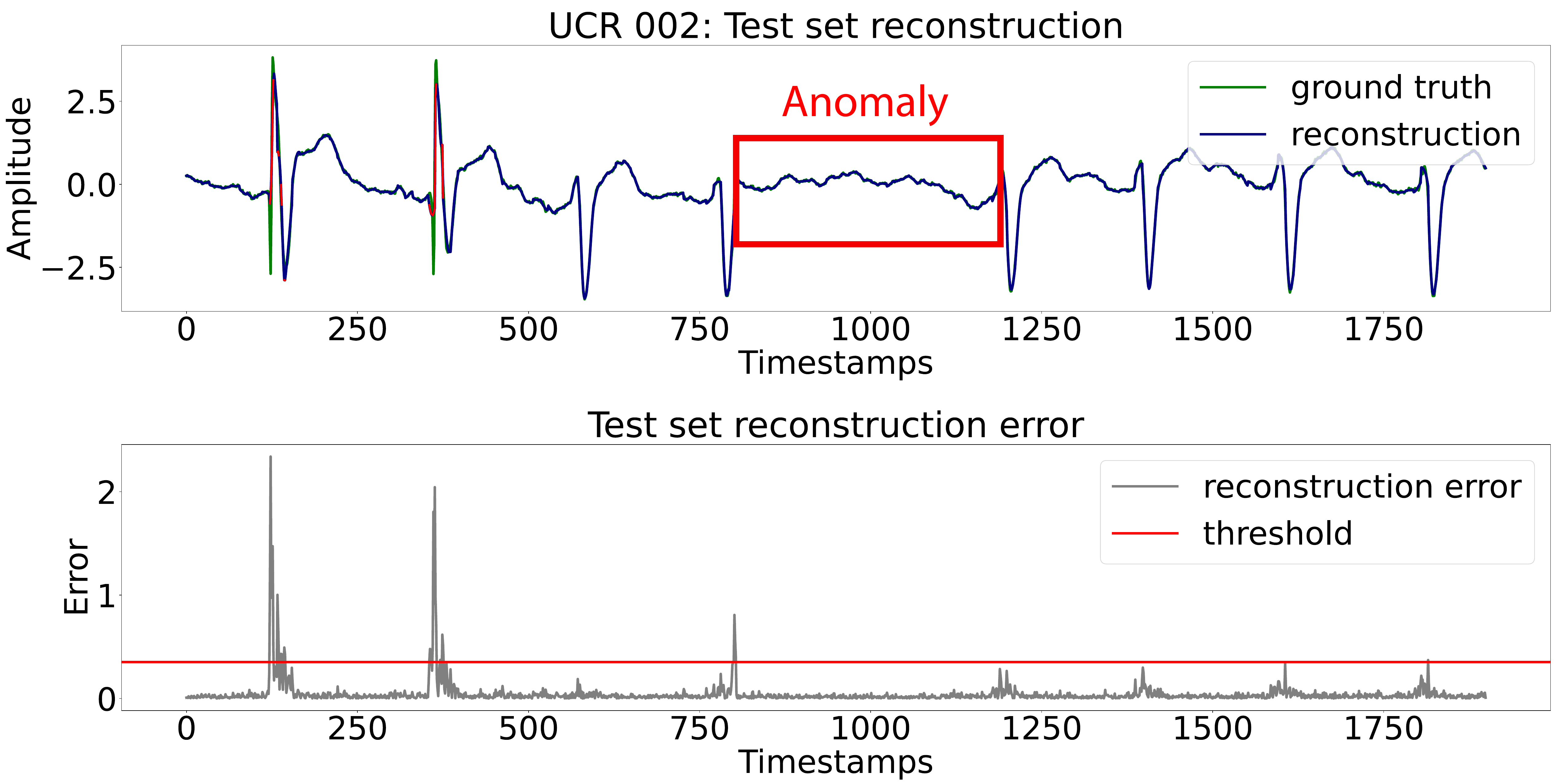}
  \caption{Detection results of LSTM-AE on a UCR test set.}
  \label{fig:lstm}
\end{figure}

Another significant challenge in time series anomaly detection lies in the varying lengths of anomaly events, which can manifest as point anomalies or sequential anomalies. To overcome the constraint of fixed detection lengths, a viable solution is the prediction-based \cite{9338317, deng2021graph} or reconstruction-based \cite{you2022semi, kieu2019outlier} model, where the value of data points are predicted, and significant deviations from these predictions are considered potential anomalies. One of the most reliable benchmark methods as discussed in a recent study \cite{kim2022towards} is to use LSTM-AE to reconstruct the temporal patterns of input data. However, as shown in Fig.~\ref{fig:lstm}, the model's strong robustness can lead it to a similarly good reconstruction of anomaly patterns as normal patterns if the anomaly sequence is continuous. In such cases, the model's predictions might accurately fit the anomaly patterns, making it challenging to maximize prediction errors between normal data points and anomalies. Additionally, these methods face sensitivity issues with the threshold, as they rely on the prediction errors. Another approach lies in discord discovery algorithms \cite{nakamura2020merlin, nakamura2023merlin}, which leverage pairwise similarity comparison to detect anomalies without necessarily relying on explicit anomaly labels and constraints of anomaly length. However, the pairwise similarity comparison process introduces more computation latency as the datasets grow larger, which poses difficulty when dealing with real-time anomaly detection.

Furthermore, some widely employed benchmark datasets (e.g., Yahoo and NASA, etc.) in previous TSAD works have been found to suffer from flaws like unrealistic anomaly density, incorrectly labeled ground truth, or instances where anomalies are so obvious that they can be easily resolved with a straightforward `one-liner' approach \cite{wu2021current, Schmidlvldb}. Additionally, widely utilized evaluation metrics, though common, have been criticized for the ill-posed point adjustments (PA) before scoring, leading to misleading impressions of model performance \cite{kim2022towards, 10.1145/3534678.3539339}. 

In light of these challenges and the presence of flawed benchmark datasets, there have been recent doubts and debates about the true capabilities of AI-based models, which we will also discuss in Section \ref{subsec:flaws}. Fortunately, this climate has provided an opportunity for the emergence of more robust deep learning models, capable of undergoing thorough testing on rigorous datasets and assessment metrics, thereby revealing their true capabilities. In this paper, we review the problem of existing benchmarks and propose a novel self-supervised learning based \textbf{\underline{Tri}}-domain \textbf{\underline{A}}nomaly \textbf{\underline{D}}etector (\textbf{TriAD}), which eliminates the necessity for anomaly labels and relieves the constraints given by anomaly lengths and shapes. 

In the case of industrial time series data, the sequences are often univariate and captured by single sensors, making it difficult to extract features from the one-dimensional data. Recognizing that anomalies can present in various ways, such as sudden changes in frequencies, shifts in residual scales, or more pronounced transformations in temporal shapes, the TriAD framework is designed to address these multifaceted characteristics. TriAD captures features across three distinct domains: temporal, frequency, and residual. By employing individual encoders for each domain, it learns unique representations that enable a comprehensive understanding of the data's underlying patterns.

To solve the challenge of label scarcity, TriAD employs data augmentations by modifying a random segment of time series input, rendering it more `abnormal'. The framework incorporates both inter-domain and intra-domain contrastive loss, which distinguishes TriAD from traditional single-domain contrastive learning methods. On the one hand, the intra-domain contrast compels original time series windows to learn common attributes while differentiating them from the augmentations. On the other hand, inter-domain contrast ensures that the learned representations from different domains are dissimilar to each other, guaranteeing that each domain develops its own unique representations. Importantly, the entire training process relies solely on normal data, thus eliminating the need for labeled anomalies.

To address the challenge of anomalies with diverse lengths, TriAD combines the strengths of both deep learning-based methods and search algorithms. The self-supervised learning framework identifies candidate windows suspected of containing anomalies and then employs a discord discovery algorithm to probe around these suspicious windows. This process enables the localization of potential anomalies of different length options. Finally, a voting system sums up and assigns scores to each data point of the test sequence, furnishing the prediction results without human intervention. The effectiveness of TriAD is further highlighted by its capability to pinpoint the potential locations of anomalies, allowing the search algorithm to concentrate exclusively on these areas. As illustrated in Section \ref{sec:merlin}, TriAD significantly reduces anomaly search window length by around 20 times as compared to state-of-the-art (SOTA) discord discovery algorithms, 
%Additionally, the search time is reduced on average by 20 times
which represents a substantial improvement in anomaly searching efficiency.

Given the complexities associated with flawed benchmark datasets and ill-posed evaluation metrics, TriAD was evaluated using the UCR Time Series Anomaly Archive \cite{wu2021current}, a carefully designed resource that aims to overcome the shortcomings of existing anomaly detection datasets. Paired with rigorous evaluation metrics (i.e., PA\%K and affiliation) that have been recently proposed \cite{kim2022towards, 10.1145/3534678.3539339}, this evaluation process ensures a transparent assessment of TriAD's capabilities, yielding a clear and credible understanding of its effectiveness.

The primary contributions of this study can be outlined as:
\begin{itemize}
\item To the best of our knowledge, this study is the first attempt to reevaluate the true potential of deep learning methods in time series anomaly detection, utilizing both rigorously designed benchmark datasets and evaluation metrics.
\item The novel approach we propose offers a robust and label-free solution for detecting anomalies of varying shapes and lengths, addressing one main challenge that plagued existing deep-learning-based methods.
% have struggled.
% to address.
\item From our experimental results, TriAD has achieved performance that is three times better than that of SOTA deep learning models, as measured by PA\%K based F1 scores. Furthermore, the accuracy of anomaly detection has improved by 50\%, and the inference time has been reduced to one-tenth compared to SOTA discord discovery algorithms.
\end{itemize}

\section{Preliminary}
In this section, we start by defining time series anomalies and introducing the frequency features used in this study. Subsequently, we discuss the problems posed by imperfect datasets and inappropriate evaluation metrics.
\subsection{Problem definition}
\begin{definition}
\textbf{Time series anomaly} Given a periodic time series $X = \{x_0, x_1, ..., x_{N-1}\}$ of length $N$, the structural modeling \cite{lai2021revisiting, tfad2022} of the sequence can be expressed as:

\begin{equation}\label{eq1}
X =\sum_{n=0}^{N-1}\left[a_n \cos \left(\omega_n n x\right)+b_n \sin \left(\omega_n n x\right)\right] + \tau\left(n\right),
\end{equation}
where the sum of sinusoidal functions characterizes the periodic or cyclical behavior of the time series, with their frequency components denoted by $\omega_n$. The terms $a_n$ and $b_n$ represent the coefficients that determine the contribution of each frequency component $\omega_n$ to the overall representation of the time series. Meanwhile, the term $\tau(n)$ accounts for non-periodic trends or unexplained variations.

Industrial sensor data typically exhibits periodic patterns due to the operational cycles of equipment \cite{9346307}. Given this characteristic, this study focuses on the important features including seasonality, shapelets, and residuals of the time series data. Consider the subsequence $X_{i, l}=\left\{x_i, x_{i+1}, \ldots, x_{i+l-1}\right\}$ of length $l$ beginning at timestamp $i$, it can be decomposed into components of seasonality $\omega$, shapelets $\rho$, and residuals $\tau$. This subsequence is considered as an anomaly of $X$ if any of the similarity measurements $\mathrm{sim}(\omega, \hat{\omega}), \mathrm{sim}(\rho, \hat{\rho})$, or $\mathrm{sim}(\tau, \hat{\tau})$ surpass a given threshold $\sigma_\omega$, $\sigma_\rho$, and $\sigma_\tau$ respectively. Here, $\hat{\omega}$, $\hat{\rho}$, and $\hat{\tau}$ refer to the expected normal behaviors of the corresponding components.
\end{definition}

\begin{table}[hbt!]
\centering
  \begin{tabular}{cc}
    \toprule
    \textbf{Quantity} & \textbf{Equations} \\
    \midrule
    Spectral amplitude & $\mathrm{A}\left(X[k]\right) = \sqrt{\mathrm{Re}\left(X[k]\right)^2+\mathrm{Im}\left(X[k]\right)^2}$ \\[5pt]
    Spectral phase & $\varphi\left(X[k]\right) = \mathrm{arctan}\frac{\mathrm{Re}\left(X[k]\right)}{\mathrm{Im}\left(X[k]\right)}$\\[5pt]
    Spectral power & $\mathrm{P}\left(X[k]\right) = \mathrm{Re}\left(X[k]\right)^2+\mathrm{Im}\left(X[k]\right)^2$\\
    \bottomrule
    \\ [-7pt]
    \multicolumn{2}{l}{\footnotesize * $\mathrm{Re} \left(X[k]\right)$ denotes the real part of $X[k]$} \\
    \multicolumn{2}{l}{\footnotesize * $\mathrm{Im} \left(X[k]\right)$ denotes the imaginary part of $X[k]$}
  \end{tabular}
\captionsetup{justification=centering}
\caption{Handcrafted frequency domain feature sets.}
\label{tab:freq_features}
\end{table}

\begin{definition}\label{freq_features}
\textbf{Time series frequency features} The Fourier transform is commonly employed to convert time series from its temporal representation into the frequency domain, revealing the power distribution across various frequencies. This process can be represented as:
\begin{equation}\label{eq1}
X[k]=\sum_{n=0}^{N-1} x_n e^{-k \frac{2\pi ni}{N}}.
\end{equation}

Here, $X[k]$ denotes the amplitude of the $k$-th harmonic, where $k$ varies from $0$ to $N-1$. In this study, we focus on three key features in the frequency domain: amplitude, phase, and power of each frequency component. These features are detailed in Table \ref{tab:freq_features}.
\end{definition}

\subsection{The `anomaly' in dataset and evaluation metrics}
\label{subsec:flaws}
Recent evaluations of time series anomaly detection have mostly relied on widely recognized datasets, such as NASA, Yahoo, and Numenta. Nonetheless, research \cite{wu2021current, lai2021revisiting} indicates that many instances within these datasets suffer from problems of anomaly mislabelling, triviality, biased distribution, and unrealistic densities. Additionally, many studies incorporate a PA process which can potentially overstate model effectiveness. This process considers an entire anomaly segment as correctly identified even if only a single instance within it is flagged as anomalous. Hence, such a methodology can lead to data leakage during evaluations since it requires tapping into the test labels to adjust predictions.

\begin{table}[hbt!]
\renewcommand{\arraystretch}{0.8}
\centering
\setlength\extrarowheight{3pt}
  \begin{tabular}{ccccc}
    \toprule
    \textbf{Dataset} & \textbf{Model} & \textbf{F1(PW)} & \textbf{F1(PA)} & \textbf{F1(PA\%K)}\\
    \midrule
    \multirow{2}{*}{\centering KPI} & LSTM-AE (Random) & \textbf{0.229} & 0.463 & \textbf{0.294}\\
     & LSTM-AE (Trained) & 0.212 & \textbf{0.524} & 0.279\\[1pt]
     \hline& \\[-2ex]
     \multirow{2}{*}{\centering SWaT} & LSTM-AE (Random) & \textbf{0.756} & 0.903 & \textbf{0.859}\\
     & LSTM-AE (Trained) & 0.454 & \textbf{0.920} & 0.537 \\[1pt]
     \hline& \\[-2ex]
    \multirow{2}{*}{\centering UCR} & LSTM-AE (Random) & 0.016 & 0.122 & 0.025\\
     & LSTM-AE (Trained) & \textbf{0.028} & \textbf{0.296} & \textbf{0.045}\\
    \bottomrule
  \end{tabular}
\captionsetup{justification=centering}
\caption{Evaluation results under new evaluation protocol.}
\label{tab:dataset}
\end{table}

\begin{figure}[hbt!]
  \centering
  \captionsetup{justification=centering}
  \includegraphics[width= 0.8\linewidth]{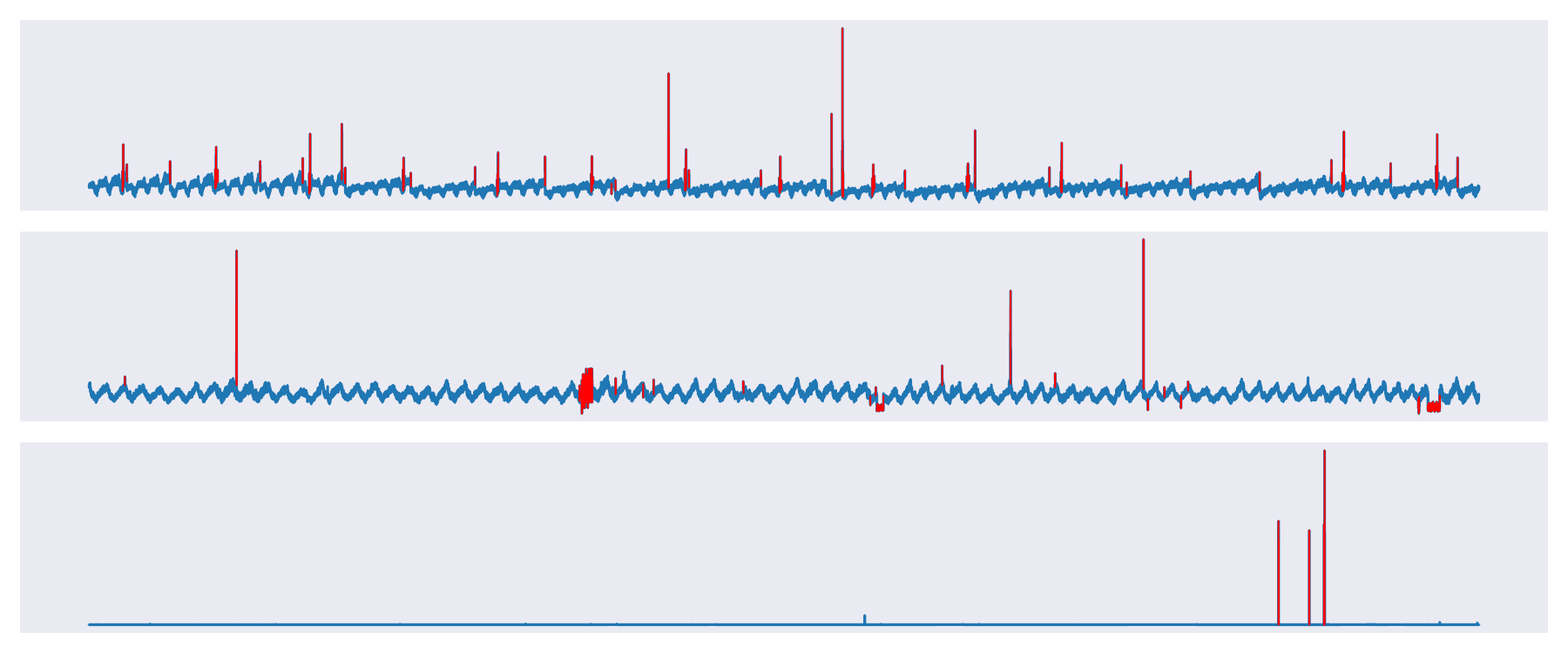}
  \caption{Time series anomalies in KPI dataset.}
  \label{fig:kpi}
\end{figure}

\begin{figure*}[hbt!]
\vspace{-0.3cm}
  \centering
  \captionsetup{justification=centering}
  \includegraphics[width= \linewidth]{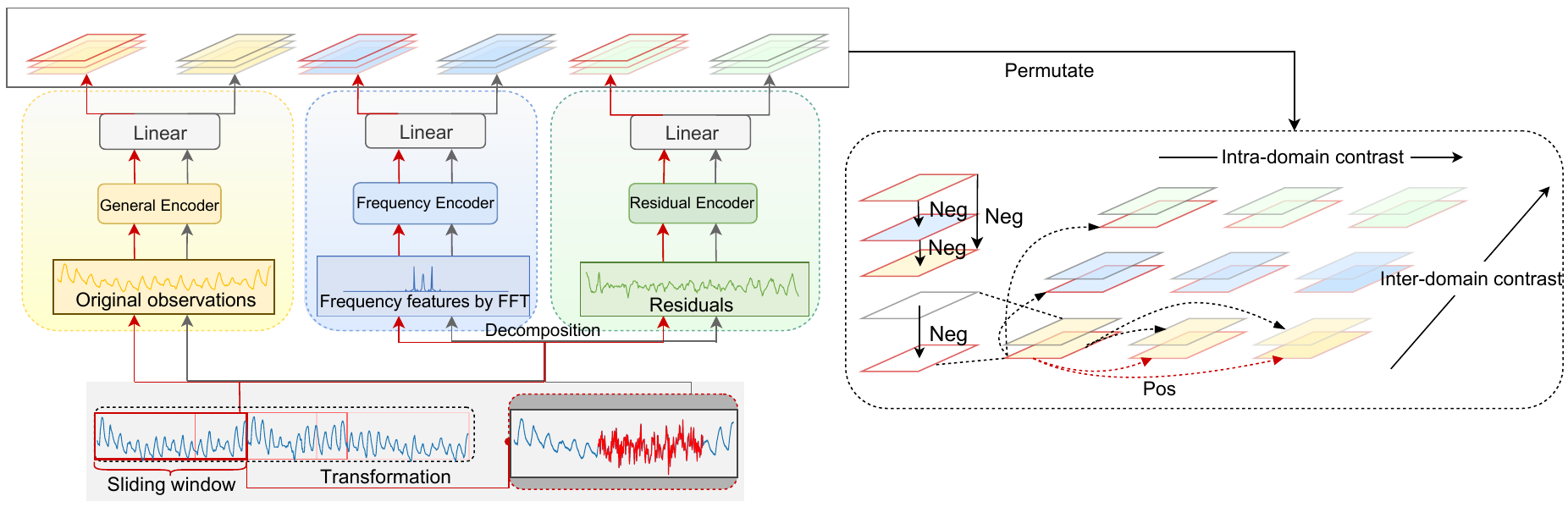}
  \caption{The architecture of TriAD.}
  \vspace{-0.5cm}
  \label{fig:framework}
\end{figure*}

Therefore, following the recently proposed evaluation protocol \cite{kim2022towards}, we examine the problem using a reliable baseline model LSTM-AE, both in its randomly initialized and trained forms on two widely used datasets, SWaT and KPI, which haven't been previously assessed. Table \ref{tab:dataset} shows a significant improvement in F1 scores when PA is employed, which potentially gives a false impression of model perfection. However, the alternative metric PA\%K which seeks to moderate the over-estimation tendencies of F1-PA and the potential underestimation of F1, reveals that a randomly initialized model might surprisingly surpass its trained performance on SWaT and KPI datasets. As depicted in Fig.~\ref{fig:kpi}, such a phenomenon is often linked to the `one-liner' issue where anomalies are so explicit that they could be more easily spotted using a single-line code setting random thresholds. The UCR dataset alleviates this problem by containing minimal trivial anomalies, as emphasized in Fig.~\ref{fig:ucr_anom}. These preliminary observations emphasize the importance of rigorous dataset selection and the utilization of reliable metrics to provide a genuine reflection of model efficacy.

\section{Methodology}
In this section, we present the proposed framework TriAD as depicted in 
Fig.~\ref{fig:framework} which captures features from three domains including temporal, frequency, and residual. This approach addresses the fundamental issue of detecting diverse anomalies which are likely to exhibit abnormality in one or more of these domains. First, we introduce the data augmentations utilized during preprocessing, followed by a discussion on the encoders that extract latent representations from the multi-domain features. Then, we explain both the inter-domain and intra-domain contrastive losses incorporated in our approach. We conclude by outlining the anomaly detection process and examining its time complexities.

\begin{figure}[hbt!]
    \centering
     \begin{subfigure}[b]{\columnwidth}
         \centering
         \includegraphics[width=\linewidth]{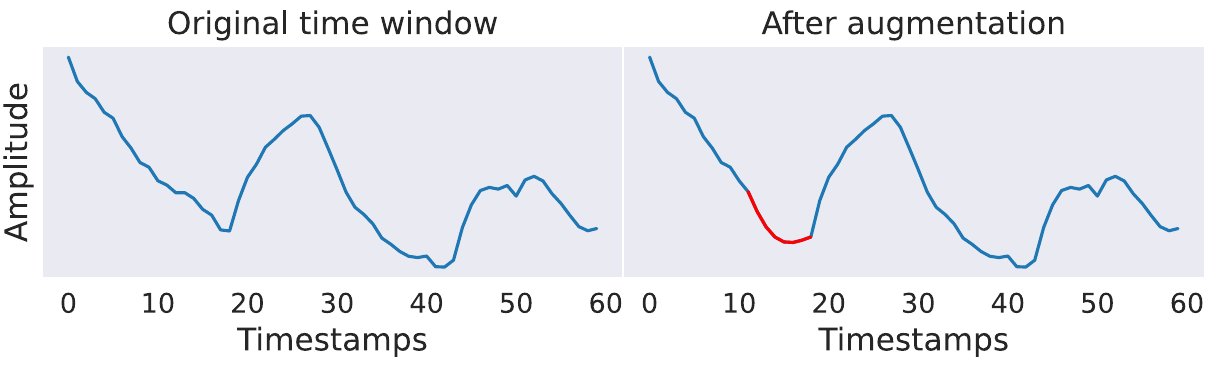}
         \caption{Magnitude warping.}
     \end{subfigure}
     \hfill
     \begin{subfigure}[b]{\columnwidth}
         \centering
         \includegraphics[width=\linewidth]{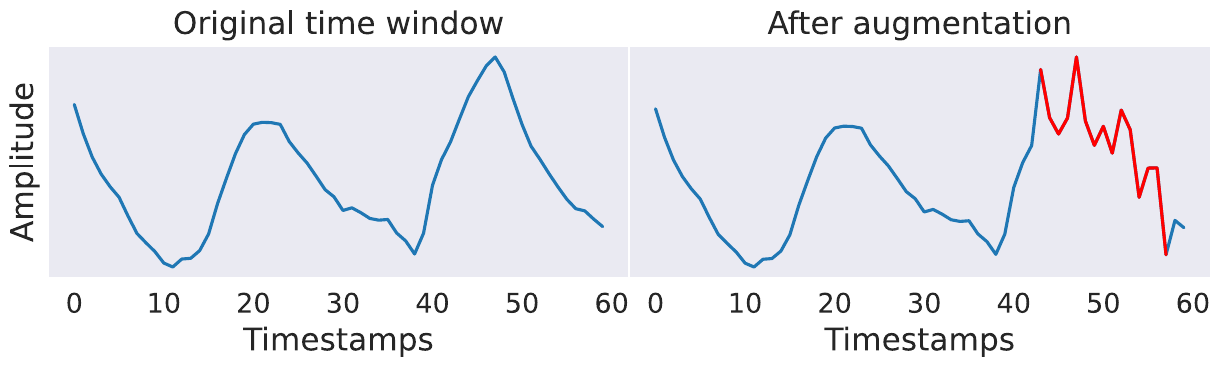}
         \caption{Jittering.}
         \label{fig:time2015}
     \end{subfigure}
\caption{Data augmentation examples.}
        \label{fig:augmentation}
\end{figure}

\subsection{Data augmentation}
Data augmentation functions by introducing diverse variations to enhance the training data, which helps strengthen model generalization and mitigate overfitting. Although data augmentation is well-defined and widely adopted in neural network-based computer vision tasks, its adoption in time series anomaly detection remains less standardized. Many strategies employed in time series augmentation draw inspiration from computer vision, incorporating methods such as adding random noise, cropping, shuffling, and scaling into the time series domain \cite{ijcai2021p631, Iwana2021survey}.

Rather than applying augmentation techniques on the entire training datasets or sliding windows to increase data volume, our approach randomly modifies a portion of input data of varying lengths, locations, and shapes. These alterations simulate potential anomalies in a time series sequence. By pairing these manipulated sequences with the original ones for the subsequent contrastive learning, the model is compelled to distinguish between sequences containing synthetically anomalous patterns and normal sequences.

To implement this, we segment the full training time series $X$ into window slices of length $L$, where the window slice starting from timestamp $i$ is denoted as $X_{i, L}$. Each window then undergoes random segment alteration using established data augmentation techniques like jittering and warping, as illustrated in Fig.~\ref{fig:augmentation}. Specifically, jittering \cite{Iwana2021survey} introduces noise into the data to obtain new samples. This process is represented as:
\begin{equation}
X_{j,l}^{\prime}(\epsilon)=\left\{x_j+\epsilon_j, x_{j+1}+\epsilon_{j+1},..., x_{j+l-1}+\epsilon_{j+l-1}\right\},
\end{equation}
where $\epsilon$ denotes the random noise added to specific timestamps in the time series data, $j$ denotes the random beginning timestamp of the augmentation, and $l$ indicates the random length of the augmentation. Meanwhile, warping adjusts a signal's magnitude based on a smooth curve. This process is achieved by a Butterworth filter \cite{mahato2022detecting} applied to the original time series, producing a filtered version emphasizing the primary frequencies of input slices. The procedure is described as:
\begin{equation}
X_{j,l}^{\prime}(\omega_c)= \mathrm{ButterworthFilter} \left(\omega_c, X_{j,l}\right),
\end{equation}
where $\omega_c$ denotes the frequency cut-off. The augmented window slice $X_{i,L}^{\prime}$ is constructed by integrating the augmented segment $X_{j,l}^{\prime}(\epsilon)$ or $X_{j,l}^{\prime}(\omega_c)$ into the original window slice $X_{i,L}$.

\subsection{Tri-domain encoder}
From the derived window slices and their respective augmentations of the time series, we proceed to extract three distinct feature representations: the temporal features which are directly represented by the univariate raw input; the frequency features which are achieved through Fourier Transform, as previously detailed in Definition \ref{freq_features}; and the residual features which are derived by eliminating the underlying periodic trends from the original input. During representation learning, each of these feature types is processed through specialized encoders. The rationale is straightforward: distinct feature spaces inherently capture varied characteristics of the data. Specifically, the temporal encoder focuses on time-evolving patterns, the frequency encoder examines spectral content and its distribution, and the residual encoder emphasizes deviations from established periodic behaviors.

In this study, we leverage the dilated convolutions and residual connections to process the time series input, capturing both short-term and long-term patterns effectively. Each of these residual blocks is structured with a pair of consecutive convolutional layers, employing same padding to ensure the preservation of spatial dimensions during the convolution operation. As we progress through the blocks, the dilation rate doubles to make the network incrementally capture wider patterns in the input sequence. Following the encoder phase, the latent representations are funneled into two shared dense layers.

For a given window slice $X_{i,L}$, we extract features represented as $X_f \in \mathbb{R}^{L \times C}$, where $C$ indicates the number of input channels (1 channel for temporal and residual features, and 3 channels for frequency features). The same-padding convolutional layer guarantees that the hidden representation learned from the stacked residual blocks adopts the form $\mathbf{h} \in \mathbb{R}^{L \times h_d}$. The impact of the dimension $h_d$ on the hidden representation is discussed in Section \ref{sec:param}. Subsequently, the dense layer compresses the high-dimensional representation into a one-dimensional output $\mathbf{r} \in\mathbb{R} ^{L}$, facilitating the similarity comparison in the contrastive learning phase. We represent the outputs of the normal window $X_{i,L}$ and its augmented counterpart $X_{i,L}^{\prime}$ derived from features of domain $d \in D$ as $\mathbf{r}_{i,d}$ and $\mathbf{r}_{i,d}^{\prime}$ respectively. 

\subsection{Tri-domain contrastive learning}
Contrastive learning is a widely used self-supervised learning technique that differentiates between positive and negative data pairs without explicit labels. Previous works in time series have employed various methods to select positive pairs, such as choosing adjacent segments \cite{tonekaboni2021unsupervised} or augmented segments \cite{ijcai2021p324}. However, these strategies aren't always effective for anomaly detection in time series. The distortions introduced by augmentation could highlight potential anomalous patterns, undermining the model's ability to detect genuine anomalies. Moreover, when anomalies are inherent in datasets, incorporating these into subseries can result in misaligned positive pairings. To overcome these unique challenges of anomaly detection in time series, we propose an innovative pairing strategy based on representations both within and across various feature domains. For intra-domain contrastive learning, we configure original and augmented samples as negative pairs, simultaneously pairing original samples from the same batch as positive pairs. On the other hand, inter-domain contrastive learning aims to maintain positive pairs as representations sourced from the same domain, while distancing representations from distinct domains to establish negative pairings. Loss functions are utilized for both the intra-domain and inter-domain contrastive learning procedures. 

\subsubsection{Intra-domain contrastive loss} 
The intra-domain contrast aims to strengthen the encoders' ability to affirm the normal patterns shared among unaltered data (through positive pairs) while also ensuring its sensitivity to distortions (through negative pairs). Given the output $\mathbf{r}_{i,d}$ and  $\mathbf{r}_{i,d}^{\prime}$ produced by the tri-domain encoder, the intra-domain contrastive loss can be formalized as:
\begin{equation}
\small
\begin{gathered}
\mathrm{sim}\left(\mathbf{r}_{i, d}, \mathbf{r}_{i, d}^{+} \right) = \sum_{j \in B} \mathbbm{1}_{\left[i \neq j\right]} \exp \left(\mathbf{r}_{i, d} \cdot \mathbf{r}_{j, d}\right)\\
\ell_{\mathrm{intra}}^{(i, d)} = -\log \frac{\mathrm{sim}\left(\mathbf{r}_{i, d}, \mathbf{r}_{i, d}^{+} \right)}{\mathrm{sim}\left(\mathbf{r}_{i, d}, \mathbf{r}_{i, d}^{+} \right) + \sum_{j \in B}\exp \left(\mathbf{r}_{i, d} \cdot \mathbf{r}_{j, d}^{\prime}\right)},
\end{gathered}
\end{equation}
where $r_{i, d}^{+}$ indicates the positive pairs of $r_{i, d}$, $i$ and $j$ represent different timestamps, and $B$ denotes the instances within the batch. 

\subsubsection{Inter-domain contrastive loss}
Inter-domain contrast aims to ensure the encoders learn domain-specific patterns and prevent the model from learning trivial features that appear consistently across different domains. This method is applied only to the representations learned from the original, non-augmented time series input. Thus, given the output representation $\mathbf{r}_{i,d}$, the corresponding loss function is defined as:
\begin{equation}
\small
\ell_{\mathrm{inter}}^{(i, d)}=-\log \frac{\mathrm{sim}\left(\mathbf{r}_{i, d}, \mathbf{r}_{i, d}^{+} \right)}{\mathrm{sim}\left(\mathbf{r}_{i, d}, \mathbf{r}_{i, d}^{+} \right) + \sum_{d \in D} \mathbbm{1}_{\left[d \neq d^{\prime}\right]} \exp \left(\mathbf{r}_{i, d} \cdot \mathbf{r}_{i, d^{\prime}}\right)},
\end{equation}
where $\mathbf{r}_{i, d^{\prime}}$ represents the learned representation from a different domain $d^{\prime}$ for the same window instance at timestamp $i$. 
\subsubsection{Total contrastive loss}
Finally, to obtain the total contrastive loss and balance the interaction between the intra- and inter-domain losses, we employ a parameter $\alpha$ that modulates their relative contributions:
\begin{equation}
\ell_{\mathrm{total}}^{(i, d)}= \left(\alpha \right)  \ell_{\mathrm{inter}}^{(i, d)} + (1-\alpha)\ell_{\mathrm{intra}}^{(i, d)}.
\end{equation}
A detailed analysis regarding the influence of various values of $\alpha$ on the model's performance is presented in Section $\ref{sec:param}$.

\subsection{Anomaly detection}
During training, the encoders are refined to pull normal patterns closer in the embedding space while distancing them from anomalous patterns. To detect anomalies during the inference stage, we first segment the test set (which contains anomalies) into individual windows and extract features from the three domains before feeding them to the encoders. The subsequent domain-specific representations are cross-compared to identify the top $Z$ most deviant windows within each domain. Given that each test set contains a single anomalous event of varying length, we simply set $Z=1$ for each domain. Consequently, TriAD nominates up to three potentially anomalous windows denoted as $X_{{t_0},L}$, $X_{{t_1},L}$, and $X_{{t_2},L}$ respectively, one from each domain. It's crucial to emphasize `up to three', as different domain evaluations might occasionally spotlight the same anomalous window. The rationale behind retaining the results from all three domains lies in the inherent unpredictability of anomalies: they can show up in patterns specific to any one of the three domains. For example, an anomaly characterized primarily by a short frequency shift would result in the least similarity with others when assessed against frequency embeddings. Such multi-domain similarity ranks based on the learned representations enhance the transparency and interpretability of TriAD.

\subsubsection{Window selection}
Given that the test set includes only one anomalous event, once we identify the three candidate windows, among which we suspect at least one contains the anomaly, we then narrow it down to the single window $X_{t,L}$ that most likely to contain the anomaly, where $t \in \{t_0, t_1, t_2\}$. Since the training set only consists of normal data, a direct strategy involves comparing the candidate windows against normal windows to find the most deviating window. By taking the stride to traverse the training data and capture data corresponding in length to the window under investigation, we can easily identify and exclude normal windows. This procedure ensures that the remaining window exhibits the highest dissimilarity compared to the training data.

\subsubsection{Discord discovery algorithm}
After detecting the most suspicious window, we isolate the anomalous sequences using a specialized similarity search technique: the discord discovery algorithm. Past research highlights the effectiveness of discord discovery algorithms in detecting anomalies in time series data \cite{nakamura2020merlin, nakamura2023merlin, yeh2018time, 8594908, yankov2008disk}. Their strength lies in evaluating the distance between each time series segment and its closest match, enabling precise anomaly detection based on the distance metrics. However, many of these algorithms are burdened with a quadratic time complexity of $O(N^2)$ \cite{yeh2018time, 8594908}. Among them, the Discord Range-Aware Gathering (DRAG) algorithm \cite{yankov2008disk} stands out as a leading method in discord discovery. Even though DRAG identifies anomalous segments regardless of their varying lengths and shapes, it's still hindered by a time complexity $O(Nl)$, where $l$ is the predetermined anomaly search length. Despite recent attempts to improve DRAG's efficiency by incorporating estimated discord distances \cite{nakamura2020merlin} or deploying Orchard’s indexing \cite{nakamura2023merlin} for nearest neighbor determination, these methods' time complexities still explode with increasing test set sizes.

By first restricting our focus to the single window $X_{t,L}$, we confine the search algorithm to scanning around this window by varying anomaly lengths. This approach significantly trims the search burden, in contrast to algorithms scanning the entire test set. In our study, we employ the cutting-edge MERLIN algorithm \cite{nakamura2020merlin} to detect variable-length anomalies around the identified window. The detection result from the dicord discovery algorithm is denoted as $\hat{X}_{t_\mathrm{dd},l}$. For comprehensive insights, we also investigate the time efficiency and predictive accuracy of the most recent MERLIN++ algorithm against TriAD in Section \ref{sec:merlin}.

\subsubsection{Anomaly scoring}
Discord discovery algorithms typically traverse through every possible anomaly length, resulting in potential anomalies being identified under various subsequence lengths. However, the outcome often involves a visual presentation, which requires manual review to further determine the detection results. In this study, we aim to reduce the need for human intervention by employing a point-wise scoring method, where every data point receives a score based on its placement in both the window flagged by TriAD and the subsequences recognized by the discord discovery algorithm.

Consider a data point $x_n$ at timestamp $n$. If $x_n$ lies within the window $X_{t,L}$, as identified by TriAD, it is awarded one `anomaly vote'. Additionally, if $x_n$ is part of a subsequence $\hat{X}_{t_\mathrm{dd},l}$, as flagged by the discord discovery algorithm for a particular anomaly length $l$, it receives another vote. The combined score for $x_n$ can be represented as:
\begin{equation}
\begin{gathered}
s_\mathrm{TriAD} \left(x_n\right) = \mathbbm{1} [x_n \in  X_{t,L}] \\
s_\mathrm{dd} \left(x_n\right) = \sum_l \mathbbm{1} [ x_n \in \hat{X}_{t_\mathrm{dd},l} ] \\
s\left(x_n\right) = s_\mathrm{TriAD}\left(x_n\right) + s_\mathrm{dd}\left(x_n\right). \\
\end{gathered}
\end{equation}

Finally, we employ a voting threshold to categorize a data point as an anomaly. For the sake of simplicity, in this work, we average the votes of data points that have received at least one vote. Data points surpassing this average are classified as anomalies. This modifiable threshold $\delta$ allows us to weigh recall and precision based on specific requirements. An extensive discussion on the influence of this threshold can be found in Section \ref{sec:case}. 

It is noteworthy that in the equation for calculating the anomaly score, we apply no normalization or weighting techniques to the scores produced by the TriAD-detected window and MERLIN search. Given our treatment of the TriAD-detected window as a preliminary filter for the MERLIN search, substantial weighting is less effective at this stage. However, in future research, we anticipate that an enhanced scoring function, possibly integrating normalization and more sophisticated weights, could significantly improve prediction outcomes.

\subsection{Time complexity analysis}
The computational efficiency of our proposed framework is determined by three primary stages: tri-window detection, single-window selection, and the discord discovery phase. The overall time efficiency is sensitive to the length chosen for time series segmentation. For a time series $X$ that's divided into $M$ windows, each of size $L$, the resulting embeddings generated by the encoders can be represented as $\mathbf{r} \in \mathbb{R}^{D \times M \times L}$. The time complexities of the individual stages are:
\begin{itemize}
\item The tri-window detection phase involves matrix multiplication, in which embeddings from different domains are permuted and computed together. This phase has a time complexity of $O(M \times L^2)$.
\item In single window selection, considering the worst-case scenario where the stride is set to 1, the Euclidean distance is computed between the candidate windows and the remaining $N-L+1$ segments from the training set. The associated time complexity is $O((N-L)\times L)$.
\item The application of the discord discovery algorithm benefits from a reduced scan length, centered around window length $L$. Its time complexity is thus reduced from the previous $O(Nl)$ to a more manageable $O(Ll)$. 
\end{itemize}
Taken together, the cumulative time complexity for the entire inference process is $O(M \times L^2) + O((N-L)\times L) + O(Ll)$. The primary factor influencing this complexity is the window length used during segmentation. It's worth noting that this length is significantly smaller than the previously dominant factor: the total data points in the test set $N$.

\section{Experiment}
In this section, we assess the robustness of our model under a rigorously designed, real-world time series anomaly detection dataset: the UCR Time Series Anomaly Archive. Furthermore, for a true representation of our model's capabilities, we adopt a variety of recently proposed rigorous evaluation metrics, providing transparent insight into its performance compared to the SOTA baselines. We also investigate the impact of varying model parameters and discuss the significance of different TriAD modules through an ablation study. Additionally, a detailed case study is included to illustrate the entire inference process and highlight the model's interpretability and versatility in handling diverse and subtle anomalies.

\subsection{Experimental settings}
In this section, we begin by introducing the UCR Time Series Anomaly Archive. Then, we detail the experimental settings, including data preparation, model training, and all model parameters employed in this study.

\subsubsection{Dataset}
\begin{figure}[hbt!]
  \centering
  \captionsetup{justification=centering}
  \includegraphics[width= 0.85\linewidth]{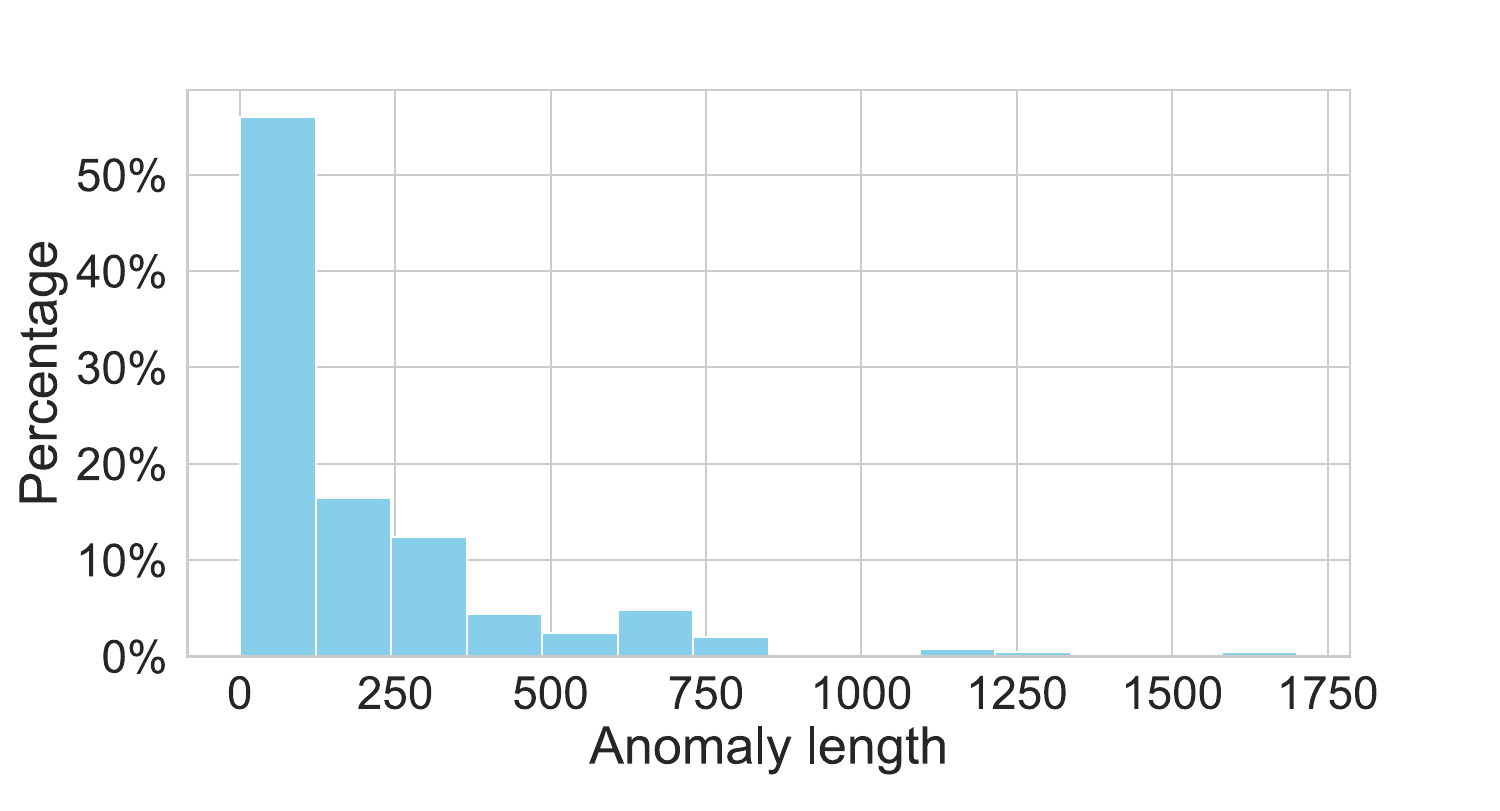}
  \caption{The length of anomalies in the datasets. }
  \label{fig:anom_len}
\end{figure}

The UCR Time Series Anomaly Archive \cite{wu2021current} is a recently introduced repository containing 250 distinct time series datasets tailored for time series anomaly detection research. Each dataset contains a single anomaly event with varying lengths, ranging from 1 to 1700. The distribution of anomaly lengths across all datasets is visualized in Fig.~\ref{fig:anom_len}, with the percentages indicating the proportion of datasets containing anomalies of a specific length. Furthermore, these datasets cover diverse domains like health, industry, and biology, exhibiting distinct anomaly types with specific characteristics. Section \ref{diversity} discusses examples of different anomaly types. Notably, only a minor portion of these datasets possess anomalies that can be addressed through the `one-liner' approach. Considering the fact that only a few real-world datasets feature explicit anomalies, these datasets ensure the generalization of the proposed anomaly detection models.

\subsubsection{Time series segmentation}
Given the diverse domains in the UCR archive, each dataset introduces its own distinct seasonal patterns. Thus, to extract pivotal features that align with the dataset's inherent seasonality and residuals, we ensure that our time series segments can capture 2.5 times the periodicity inherent to each dataset. Also, to further enhance the feature extraction process, we set a stride of a size to be a quarter of the window's length, guaranteeing sufficient coverage during time series segmentation.

\subsubsection{Model training}
For our anomaly detection study, we train distinct models for each of the 250 datasets from the UCR archive, and each model was applied to its respective test set to get detection outcomes. We employ a default batch size of 8, a learning rate set at 0.001, and we conducted training over 20 epochs for each dataset. Moreover, to ensure our models remained generalized and avoided overfitting, we use 10\% of the training data to serve as a validation set. To ensure the reliability of our results, we test the model under five distinct seeds and compute both the average and standard deviation, as presented in Table \ref{tab:deep all}.

\subsubsection{Model parameters}
Our model is carefully designed with an encoder that incorporates 6 residual blocks. This encoder produces embeddings with dimensions denoted as $h_d = 32$. To strike the right balance in the learning process, we adjusted the contrastive loss parameter $\alpha$ to 0.4. This parameter harmonizes the contributions from both intra-domain and inter-domain contrastive losses. For an in-depth exploration of the influence of key parameters on the model's performance, please refer to Section \ref{sec:param}.

\subsection{Performance comparison}
In this section, we compare the performance of our proposed method with SOTA deep learning models from recent studies using two rigorous evaluation metrics: PA\%K and affiliation metrics. For an unbiased comparison, we test each model using its source code and exclude any PA processes prior to subjecting the predictions to our redefined evaluation metrics. Further deepening our analysis, we conduct an evaluation to compare event-wise accuracy with the latest discord discovery algorithm. Our results demonstrate that our model exhibits robust performance from both point-wise and event-wise perspectives.

\begin{table*}
\renewcommand{\arraystretch}{0.8}
\centering
\setlength\extrarowheight{7pt}
  \begin{tabular}{cccccccccc}
    \toprule
     \multirow{2}{*}{\textbf{Model}} & \multirow{2}{*}{\textbf{F1(PW)}} & \multirow{2}{*}{\textbf{F1(PA)}} & \multicolumn{3}{c}{\textbf{F1(PA\%K)}} & & \multicolumn{3}{c}{\textbf{Affiliation}}\\ \cline{4-6} \cline{8-10}
    & & &  \textbf{Precision-AUC} & \textbf{Recall-AUC} & \textbf{F1-AUC} & & \textbf{Precision} & \textbf{Recall} & \textbf{F1}\\
    \midrule
    LSTM-AE (Random) & 0.016 & 0.122 & 0.023 & 0.069 & 0.025 & & 0.389 & 0.726 & 0.507\\ \hline
    LSTM-AE (Trained) & 0.028 & 0.296 & 0.031 & 0.175 & 0.045 & & 0.507 & 0.992 & 0.671\\ \hline
    USAD & 0.062 & 0.100 & 0.055 & \textbf{0.914} & 0.070 & & 0.530 & \textbf{0.999} & 0.693\\ \hline
    TS2Vec & 0.006 & 0.192 & 0.040 & 0.022 & 0.011 & & 0.376 & 0.676 & 0.483\\ \hline
    Anomaly Transformer & 0.013 & 0.247 & 0.015 & 0.082 & 0.022 & & 0.488 & 0.650 & 0.325\\ \hline
    MTGFlow & 0.058 & 0.101 & 0.054 & 0.771 & 0.070 & & 0.524 & 0.984 & 0.684\\ \hline
    DCdetector & 0.018 & \textbf{0.298} & 0.023 & 0.074 & 0.032 & & 0.484 & 0.959 & 0.643\\ \hline
    \textbf{TriAD} & \textbf{0.237}  & 0.294 & \textbf{0.233} $\pm$0.009  & 0.398 $\pm$0.015  & \textbf{0.263} $\pm$0.010 & & \textbf{0.722} $\pm$0.014 & 0.736 $\pm$0.014 & \textbf{0.729} $\pm$0.014\\
    \bottomrule
    \multicolumn{8}{l}{\footnotesize * Window-based detection accuracy of TriAD: tri-window 0.531 $\pm$ 0.017, single window 0.482 $\pm$ 0.019} \\
  \end{tabular}
\captionsetup{justification=centering}
\caption{Overall comparison with SOTA deep learning models on 250 UCR datasets.}
\label{tab:deep all}
\end{table*}

\subsubsection{Comparison with SOTA deep learning models}
In the study presented by \cite{kim2022towards}, the authors demonstrate that a randomly initialized LSTM-AE can serve as a benchmark in time series anomaly detection due to its impressive performance against many state-of-the-art models. Inspired by this study, we compare recent deep learning models against both the trained and randomly initialized LSTM-AE using the UCR dataset. It is worth noting that we've strategically chosen a rich range of baseline models, including cutting-edge techniques like attention mechanisms, autoencoders, adversarial, and self-supervised learning. Here's a concise breakdown of these models:
\begin{itemize}
\item \textbf{LSTM-AE} \cite{kim2022towards}: As a recommended baseline, this model outperforms several SOTA deep learning models. The simple architecture uses an untrained autoencoder with a single-layer LSTM. We present results for both its variants: randomly initialized and trained.
\item \textbf{USAD} \cite{usad20}: This unsupervised anomaly detection model employs two autoencoders, each comprising a shared encoder and distinct decoders. Its two-stage training involves an autoencoder phase followed by adversarial training.
\item \textbf{TS2Vec} \cite{Yue2021TS2VecTU}: This work utilizes contrastive learning hierarchically across augmented context views, facilitating robust timestamp representations. 
\item \textbf{Anomaly Transformer} \cite{xu2022anomaly}: This work introduces the Anomaly-Attention mechanism to measure association discrepancies, which are determined by comparing the prior association of each timestamp to its series association derived from the self-attention map.
\item \textbf{MTGFlow} \cite{MTGFlow23}: This method addresses time series anomaly detection using dynamic graphs and entity-aware normalizing flow. Its foundation is the assumption that abnormal events display sparser densities than normal ones.
\item \textbf{DCdetector} \cite{dcdetector23}: This work leverages a dual attention contrastive learning approach to effectively distinguish between normal and abnormal data representations. It utilizes patching-based attention networks to capture the temporal dependency in time series. 
\end{itemize}
Based on the problems associated with the PA process detailed in Section \ref{subsec:flaws}, our study adopts the optimized PA-based metric, PA\%K, and the event distance-centric method, the affiliation metrics. Here's a more in-depth look into these evaluation metrics:
\begin{itemize}
\item \textbf{PA\%K}  \cite{kim2022towards}: This metric optimizes the point-wise evaluation by considering the trade-off between conventional F1 measurements and the ill-posed PA. It measures the model performance by weighing the proportion of accurately identified anomalies. The process is expressed as:
\begin{equation}
\hat{y}_n= \begin{cases}1, & \text { if } s\left(x_n\right)>\delta \text { or } \\ & n \in \mathcal{A} \text { and } \frac{\left|\left\{n^{\prime} \mid n^{\prime}\in \mathcal{A}, s\left(x_{n^{\prime}}\right)>\delta\right\}\right|}{\left|\mathcal{A}\right|}>\mathrm{K} \\ 0, & \text { otherwise }\end{cases}
\end{equation}

In this formula, the detection output is symbolized as $\hat{Y} = {\hat{y}_0, \hat{y}_1, ..., \hat{y}_{N-1}}$. Each component is marked by 0s and 1s, denoting the anticipated anomaly labels. Meanwhile, $\mathcal{A}$ represents the anomaly sequence wherein all constituent points are labeled as 1, collectively representing a single anomaly event. In this study, we assess the F1 scores using values of $K$ ranging from 1 to 100 and determine the optimized scores using the Area under the Curve (AUC).

\item \textbf{Affiliation} \cite{10.1145/3534678.3539339}: This metric measures the performance of predictions from the event-wise perspective. It provides compensations for incorrect predictions close to the actual event and offers more rewards for long anomalous events. It then transforms the temporal distance between the actual events and predictions into a distribution function, which in turn is used to compute the precision and recall based on probabilities. Since each test set contains only one anomalous event, the formula is represented as:
\begin{equation}
\begin{gathered}
P_{\text{precision}}=\frac{1}{\left|\operatorname{\hat Y^{\prime}}\right|} \int_{\hat y^{\prime} \in \operatorname{\hat Y^{\prime}}} \bar{F}_{\text {precision }}\left(\operatorname{dist}\left(\hat y^{\prime}, \mathcal{A}\right)\right) d y^{\prime} \\
P_{\text {recall }}=\frac{1}{\left|\mathcal{A}\right|} \int_{a \in \mathcal{A}} \bar{F}_{\text { recall }}\left(\operatorname{dist}\left(a, \hat {Y^{\prime}}\right)\right) da
\end{gathered}
\end{equation}
\end{itemize}
In the above, $\hat Y^{\prime}$ denotes all predictions tagged as anomalies. The distance function $\mathrm{dist}\left(\hat y^{\prime}, \mathcal{A}\right) = \mathrm{min}_{a\in \mathcal{A}}|\hat y^{\prime}-a|$, and $\mathrm{dist}\left(a, \hat Y^{\prime}\right) = \mathrm{min}_{\hat y^{\prime} \in \hat Y^{\prime}}|a-\hat y^{\prime}|$ determine the temporal separation between the predicted and actual events. Additionally, the function $\bar{F}$ denotes the survival function derived from the distance-based cumulative distribution.

The comparison results against SOTA deep learning models are presented in Table \ref{tab:deep all}. As observed, most deep learning models outperform the benchmark model: a randomly initialized LSTM-AE. This performance is expected given the reliability of the expertly crafted anomaly types present in the dataset. When viewed from a point-wise perspective, nearly all baseline models exhibit a substantial benefit from PA, resulting in an F1 score increase ranging between approximately 2 to 30 times. However, the PA\%K adjustment offers a more realistic depiction, highlighting the true effectiveness of model performance. The results suggest that a majority of models struggle to detect a significant fraction of data points within anomalous events, and their predictive capability diminishes as $K$ grows. In contrast, TriAD maintains consistent performance both pre and post-PA. This is because TriAD is inherently designed to allocate anomalies in the most likely areas. As a result, its precision is at least 4 times greater than other methods, culminating in an F1 score enhancement of at least 4 times post-optimization. From an event-wise perspective, while other methods may demonstrate a higher recall due to falsely identifying many positive anomalies, they suffer a precision decline in compensation. In comparison, TriAD significantly reduces false positives by focusing on anomalous regions, ultimately resulting in an F1 score boost of at least 6\% under affiliation evaluation.

\subsubsection{Comparison with discord discovery algorithm} \label{sec:merlin}

\begin{table}[hbt!]
\renewcommand{\arraystretch}{0.8}
\centering
\setlength\extrarowheight{6pt}
  \begin{tabular}{cccc}
    \toprule
    \textbf{Dataset} & \textbf{Model} & \textbf{Accuracy} & \textbf{Inference Time (mins)}\\
    \midrule
    \multirow{3}{*}{\centering UCR} & Merlin++ &  0.424 & 14.5\\ \cline{2-4}
     & TriAD (tri-window) & 0.681 & 0.99\\ \cline{2-4}
     & TriAD (single window) & 0.623 & 1.01\\
    \bottomrule
  \end{tabular}
\captionsetup{justification=centering}
\caption{Comparision with MERLIN++ by 62 shortest UCR datasets.}
\label{tab:merlin++}
\end{table}

MERLIN++ is renowned for its capability to accelerate discord search time while maintaining the same accuracy level as the original MERLIN by the Orchard’s indexing algorithm. For a comparative assessment against this SOTA discord discovery algorithm, we adopt the evaluation metrics originally employed by MERLIN++. Here, accuracy is determined by the count of anomalous events successfully detected among the test set, and a prediction within a margin of 100 data points surrounding the anomaly is deemed correct. They arranged the 250 datasets in ascending order of length, and in the initial 62 datasets, MERLIN++ correctly identified 29 anomaly events. In our approach, we straightforwardly utilize the windows identified by TriAD, measuring accuracy based on their inclusion of anomalies. As illustrated in Table \ref{tab:merlin++}, when comparing with outcomes from the MERLIN++ study, we achieve an increase of approximately 50\% in accuracy. Additionally, our method is a minimum of tenfold faster than MERLIN++ during the inference phase. Note that across 250 datasets, TriAD has an average tri-window detection accuracy of 0.531 ($\pm$0.017) and a single-window detection accuracy of 0.482 ($\pm$0.019).

\begin{figure}[t!]
  \centering
\captionsetup{justification=centering}
  \includegraphics[width= \linewidth]{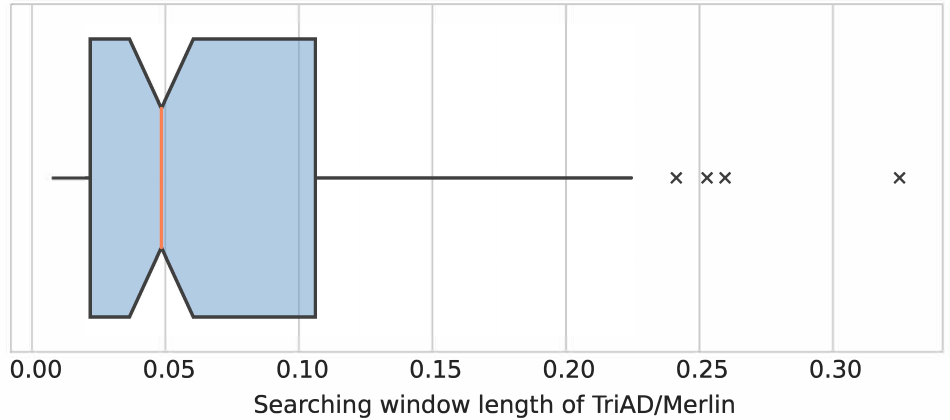}
  \caption{The ratio of anomaly search length of TriAD to MERLIN.}
  \label{fig:search_len}
\end{figure}

In the final stage of the TriAD framework, we integrate MERLIN to accurately pinpoint anomalies, offering flexibility with variable length options. This stage significantly benefits from limiting the search length to a specific window size. In this study, to unlock the full potential of MERLIN, we introduce padding with data points both before and after the chosen window, which provides added context regarding the typical appearance of normal data. As illustrated in Fig.~\ref{fig:search_len}, TriAD on average provides a search length that is 20 times shorter than that required by MERLIN across all datasets, considerably speeding up the algorithm. Furthermore, as TriAD identifies more accurate windows before the discord search phase, it potentially assists MERLIN in sidestepping false positives that originate from a vast of normal data. This focus prompts the algorithm to identify anomalies within a specified region, thereby enhancing its prediction accuracy.

\begin{figure*}[t!]
  \centering
  \captionsetup{justification=centering}
  \includegraphics[width= 0.85\linewidth]{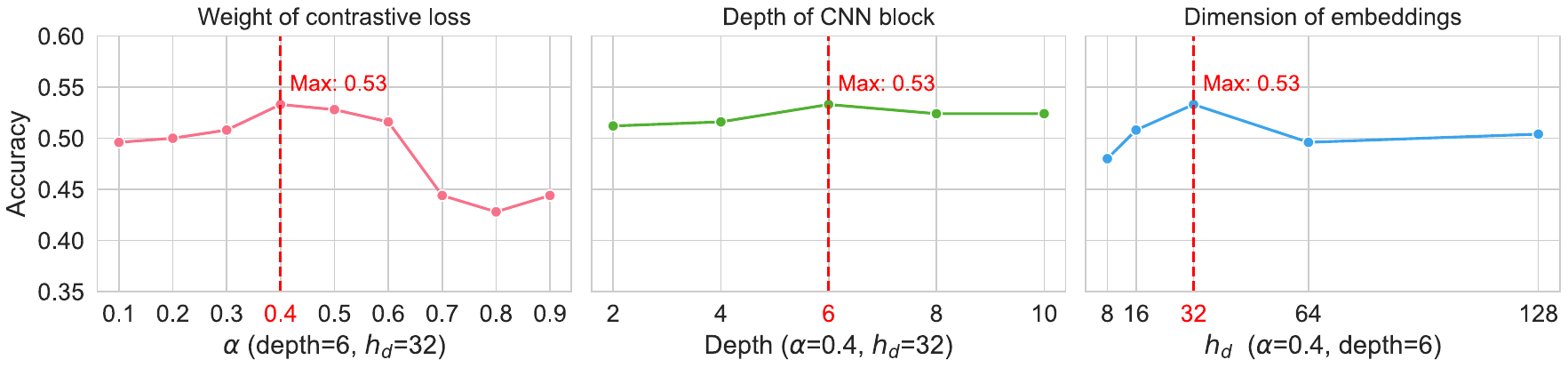}
  \caption{Parameter study of TriAD: the weight of contrastive loss ($\alpha$), the number of CNN blocks (depth), the dimension of representation learned by encoders ($h_d$).}
  \label{fig:param}
\end{figure*}

\subsection{Ablation study}
\begin{figure}[hbt!]
  \centering
  \captionsetup{justification=centering}
  \includegraphics[width= 0.9\linewidth]{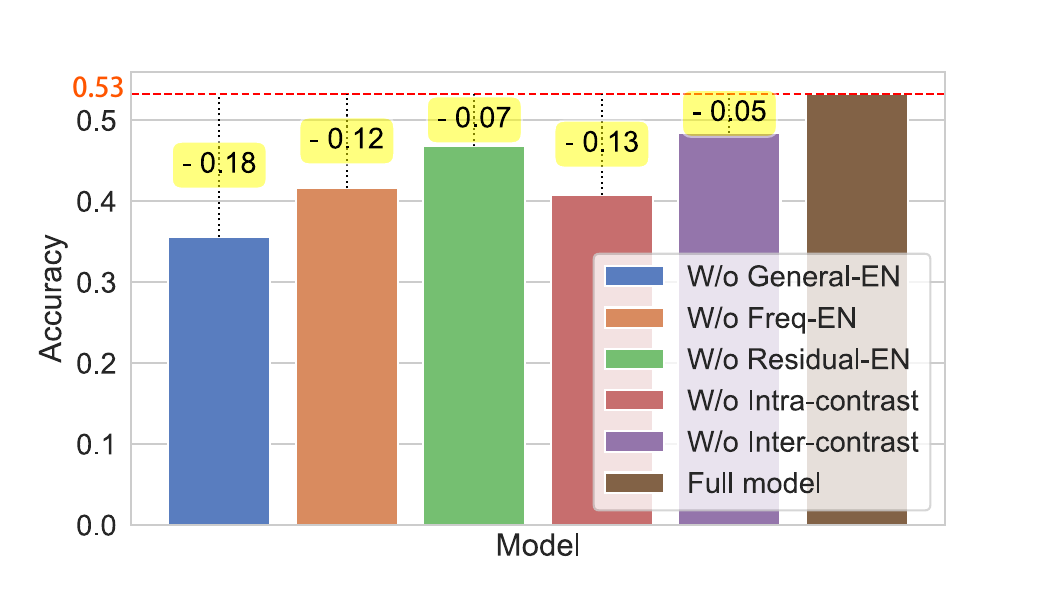}
  \caption{Ablation study of TriAD.}
  \label{fig:ablation}
\end{figure}

To understand the significance of each module within TriAD, we performed an ablation study. This experiment involves removing individual modules from the framework and observing the subsequent decline in performance. We fine-tune our model using the anomaly detection accuracy of tri-window, as this metric provides a direct measure of TriAD's performance and influences subsequent stages. As depicted in Fig.~\ref{fig:ablation}, among the three encoders in TriAD, the general and frequency encoders play a vital role in representation learning for anomaly detection, while the residual encoder holds less significance. This might be attributed to the datasets' predominate anomalies related to temporal and frequency variations, with residual shifts being less crucial. Moreover, intra-domain contrastive learning holds greater weight than its inter-domain counterpart, suggesting that data augmentation significantly aids in distinguishing multi-view anomaly representations. 

\subsection{Parameter study}\label{sec:param}
To assess the influence of various parameters in TriAD, we focused on three key parameters: the weight of the contrastive loss ($\alpha$), the number of CNN blocks (depth), and the encoder's representation dimension ($h_d$). As depicted in Fig.~\ref{fig:param}, optimal performance is attained when $\alpha$ strikes a balanced weight for multi-domain contrastive loss. Moreover, as $\alpha$ decreases, the model increasingly benefits from intra-domain contrastive learning, aligning with our insights from the ablation study. The performance remains relatively stable regardless of the number of CNN blocks, though a depth of 6 slightly edges out the rest. Lastly, a representation dimension of 32 yields the most favorable outcomes, with larger dimensions potentially leading to overfitting in the model.

\subsection{Case study: UCR ``025"}\label{sec:case}

\begin{figure}[hbt!]
  \centering
  \captionsetup{justification=centering}
  \includegraphics[width= \linewidth]{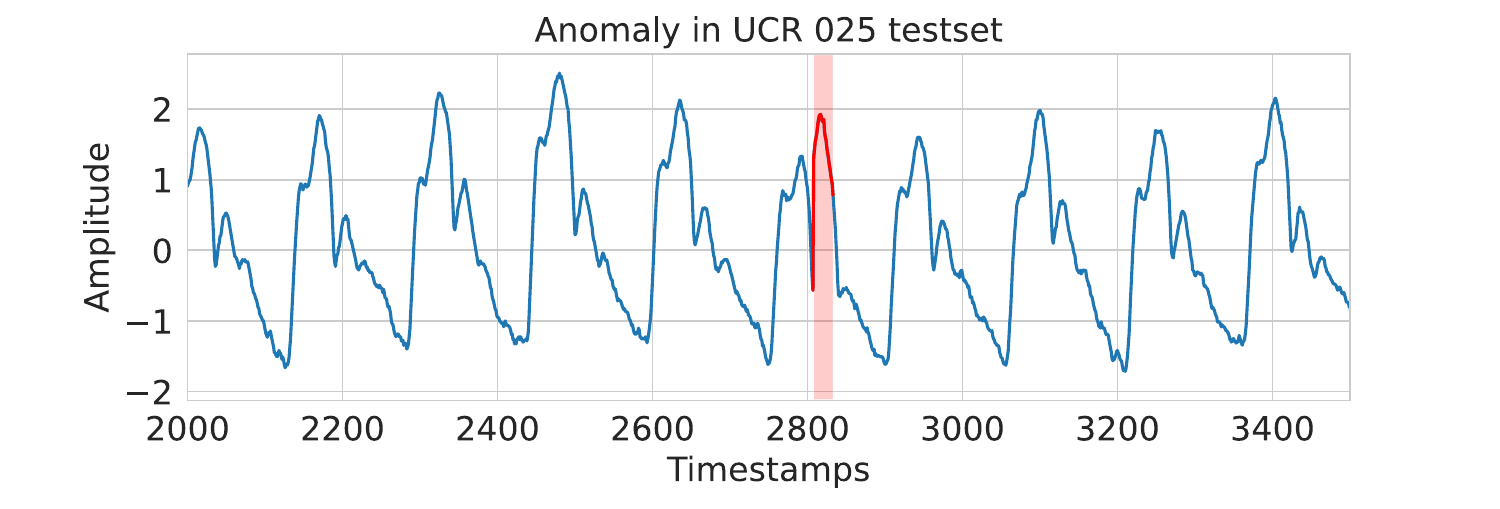}
  \caption{The anomalies in UCR 025 testset.}
  \label{fig:025_anomaly}
\end{figure}

To elucidate the anomaly detection methodology of TriAD, we present a case study using the UCR dataset ``025" with a length of 4702 data points in the test set. As illustrated in Fig.~\ref{fig:025_anomaly}, the anomaly contains 27 data points. This segment displays a subtle frequency shift compared to the norm; while normal segments have a smaller peak following the primary peak, the anomaly omits this smaller peak.

\begin{figure}[hbt!]
  \centering
  \captionsetup{justification=centering}
  \includegraphics[width= \linewidth]{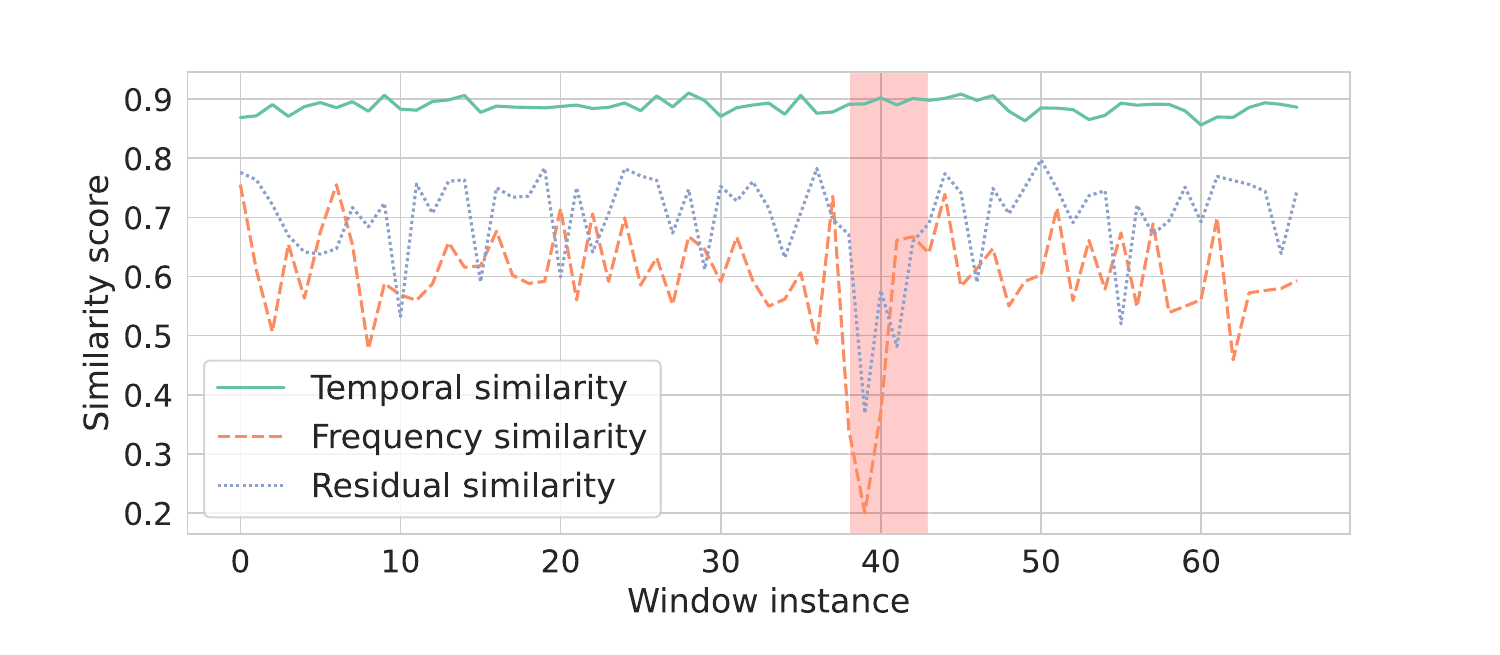}
  \caption{The window similarity scores of different domain features. }
  \label{fig:025_sim}
\end{figure}

We begin by segmenting the test set into 67 windows, with each window covering approximately 2.5 periods or around 350 data points. Each window's domain-specific encoders then learn representations and facilitate a pairwise similarity comparison amongst them, as depicted in Fig.~\ref{fig:025_sim}. Notably, the window instance at index 39 obtains the lowest similarity score in both the frequency and residual domains, distinguishing it from the rest. This result suggests that this window likely contains anomalies marked by significant shifts in frequency and residuals, even as its temporal similarities remain relatively stable and do not exhibit such drastic drops.

\begin{figure}[hbt!]
  \centering
  \captionsetup{justification=centering}
  \includegraphics[width= \linewidth]{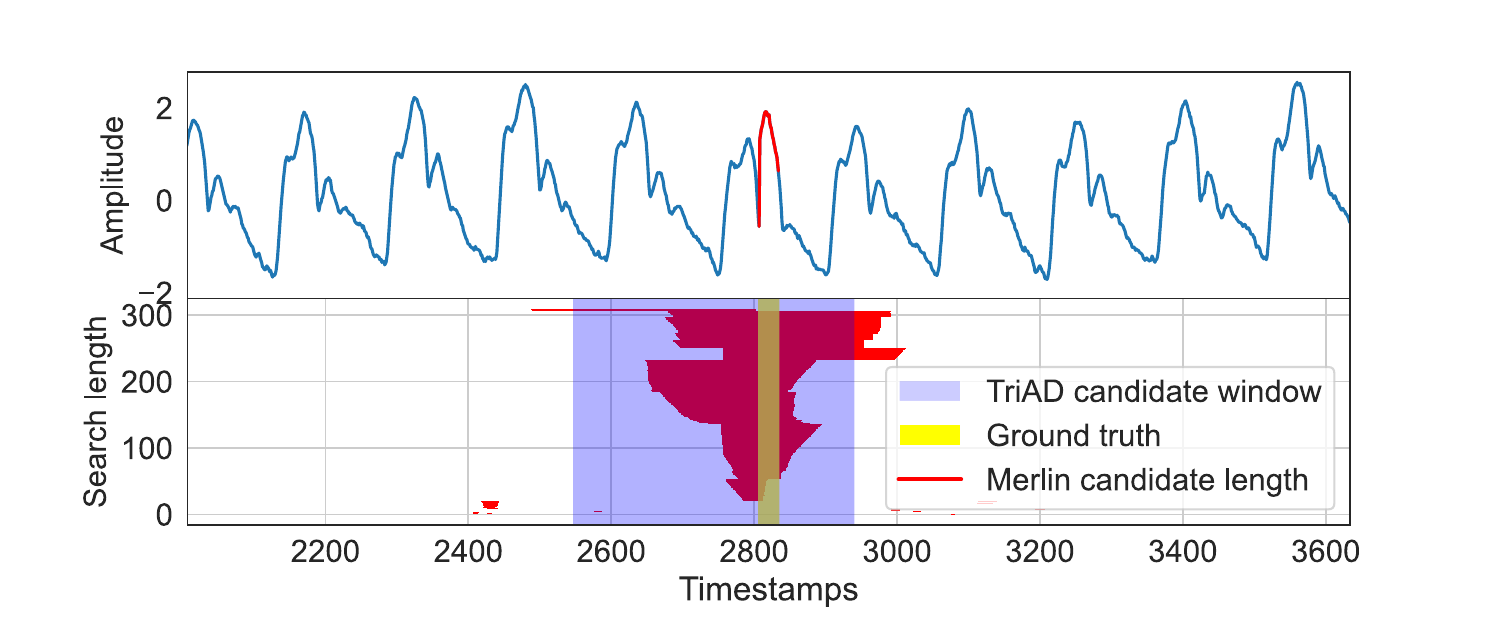}
  \caption{TriAD integrates MERLIN algorithms to find anomalies.}
  \label{fig:025_merlin}
\end{figure}

After refining our selection to identify the most suspicious window, we introduce padding before and after it, preparing it for MERLIN's analysis. As depicted in Fig.~\ref{fig:025_merlin}, with the window highlighted by TriAD, MERLIN proceeds to search within a flexible range of lengths, ranging from 3 to 300. Each search identifies time series segments of a specific length that appear most distinct from the others. The outcomes reveal that almost all results from the discord search, irrespective of their length, concentrate within the area where the anomaly resides.

\begin{figure}[hbt!]
  \centering
  \captionsetup{justification=centering}
  \includegraphics[width= \linewidth]{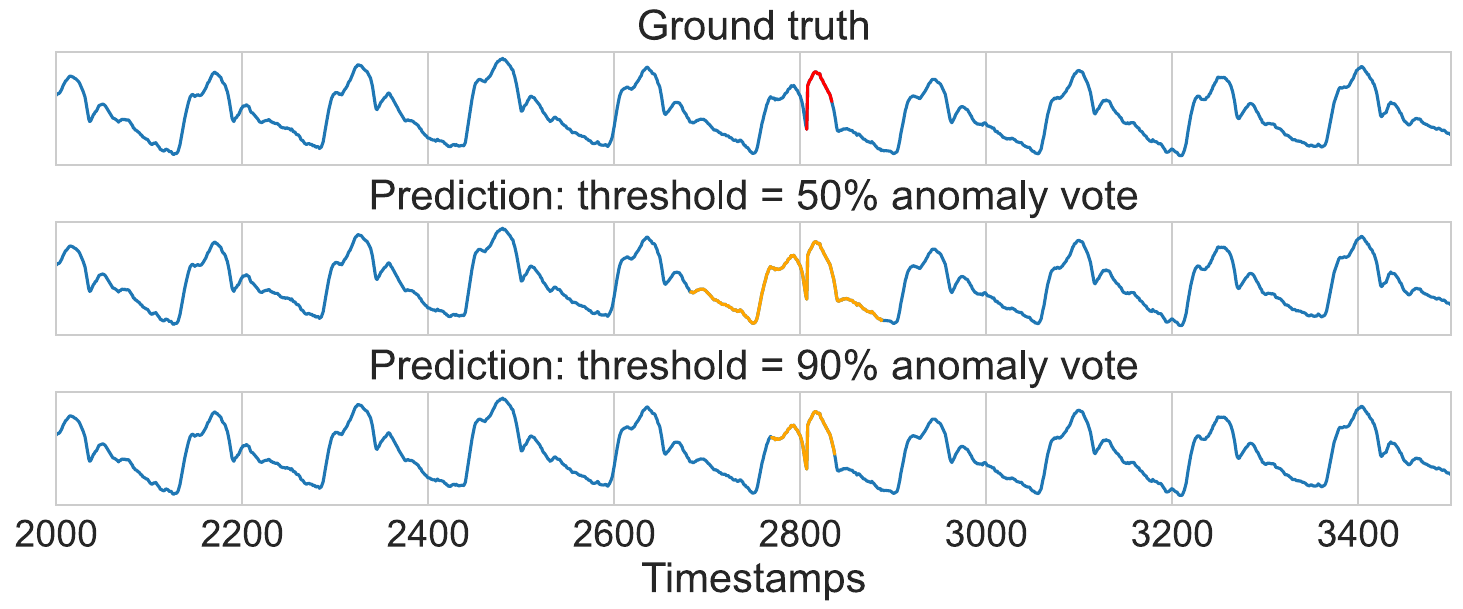}
  \caption{The anomaly detection results under different thresholds.}
  \label{fig:025_vote}
\end{figure}

While MERLIN provides a visual insight into the likely distribution of anomalies in the test set, interpreting these results requires human involvement. To address this, we convert the visual data into detection outcomes using our voting system. As previously discussed, in this work, we employ the average voting scores to determine anomaly thresholds. Additionally, we explore the effects of varying this threshold on our detection results. As illustrated in Fig.~\ref{fig:025_vote}, setting the threshold above the 90th percentile of voting scores notably improves detection precision. This increase in threshold filters out more of the normal data, preserving only the most distinct segments.

\subsection{Diversity of anomalies}\label{diversity}
\begin{figure}[hbt!]
  \centering
  \captionsetup{justification=centering}
  \includegraphics[width= \linewidth]{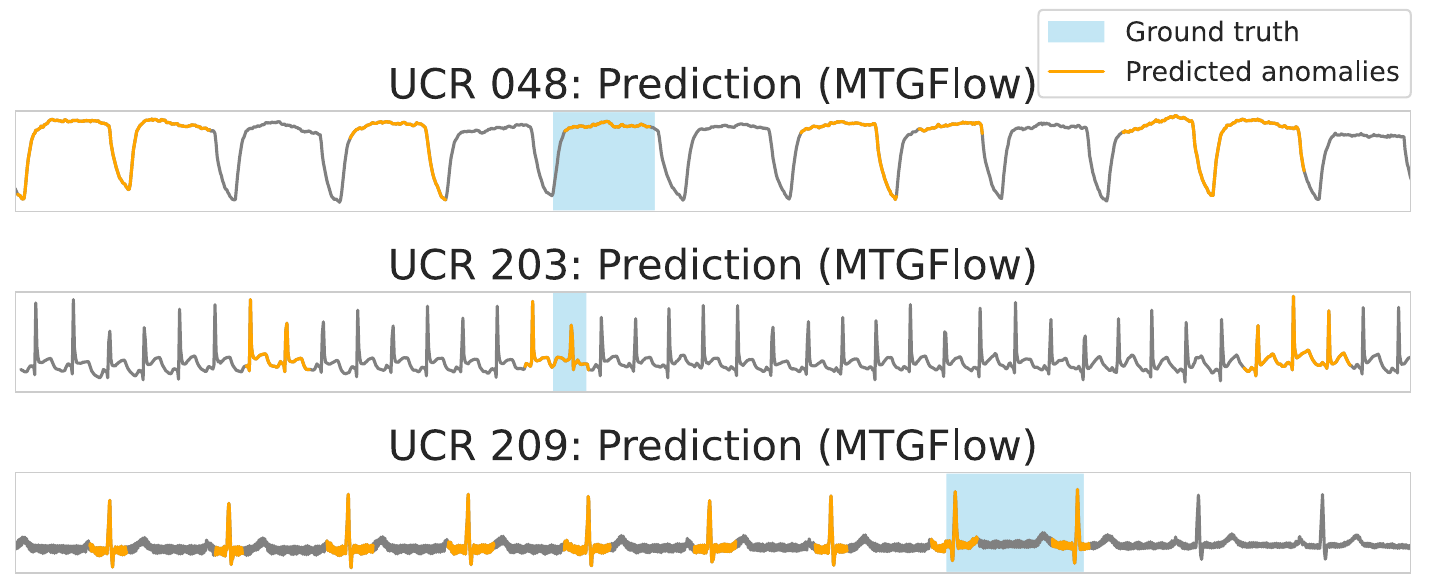}
  \caption{Anomaly detection results by MTGFlow.}
  \label{fig:mtgflow}
\end{figure}

\begin{figure*}[hbt!]
\centering
\begin{subfigure}{.325\textwidth}
    \centering
    \includegraphics[width=\linewidth]{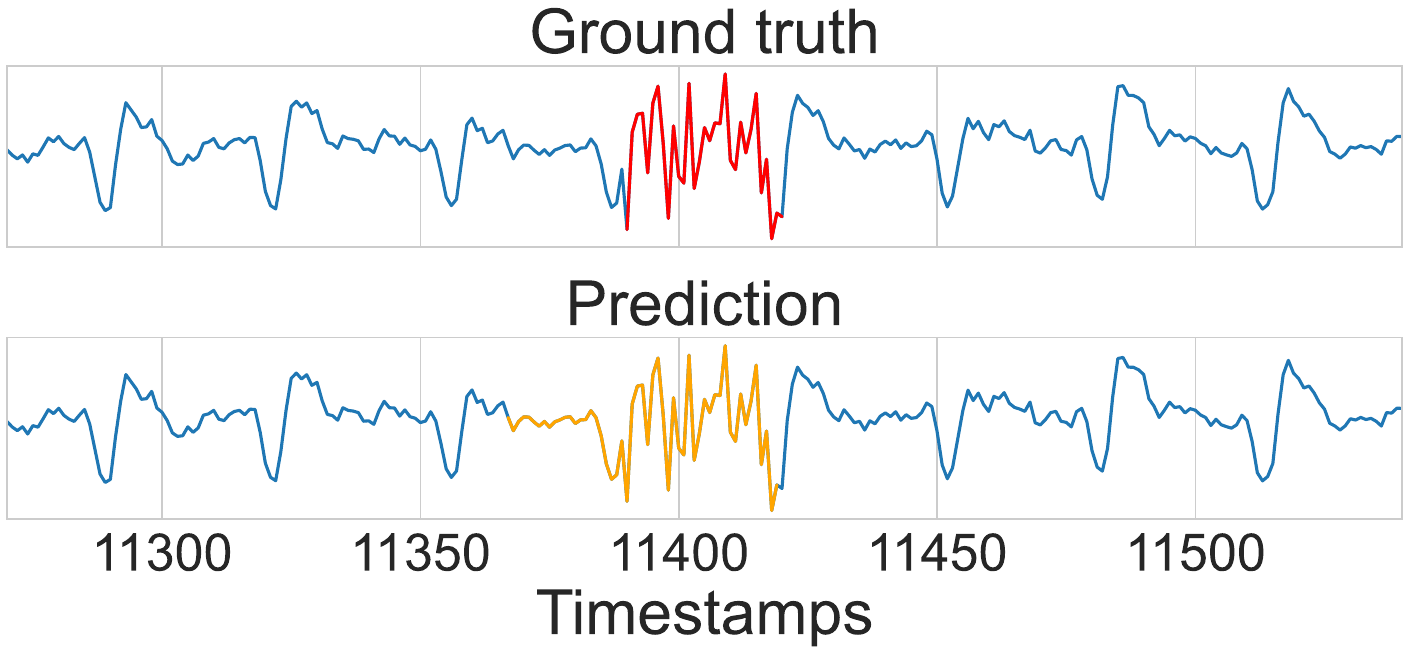}  
    \caption{UCR 039: Noise-like anomaly}
\end{subfigure}
\begin{subfigure}{.325\textwidth}
    \centering
    \includegraphics[width=\linewidth]{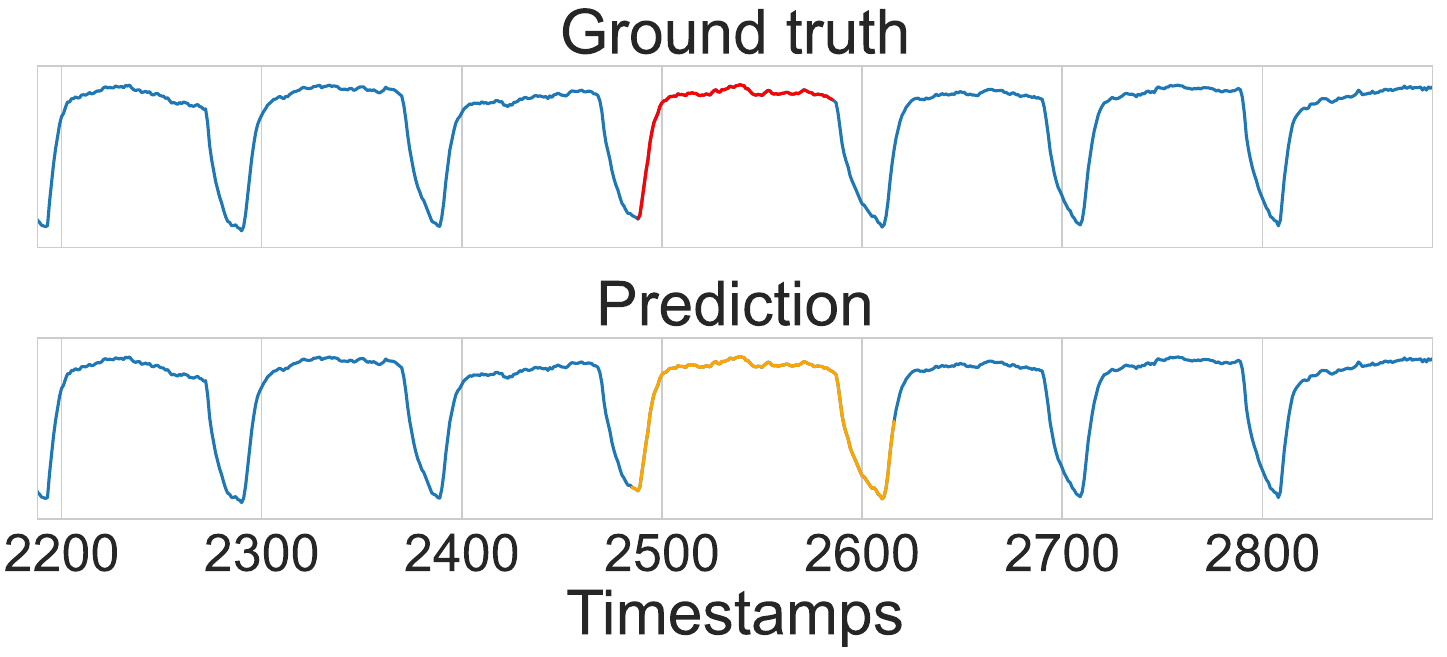}  
    \caption{UCR 048: Duration anomaly}
\end{subfigure}
\begin{subfigure}{.325\textwidth}
    \centering
    \includegraphics[width=\linewidth]{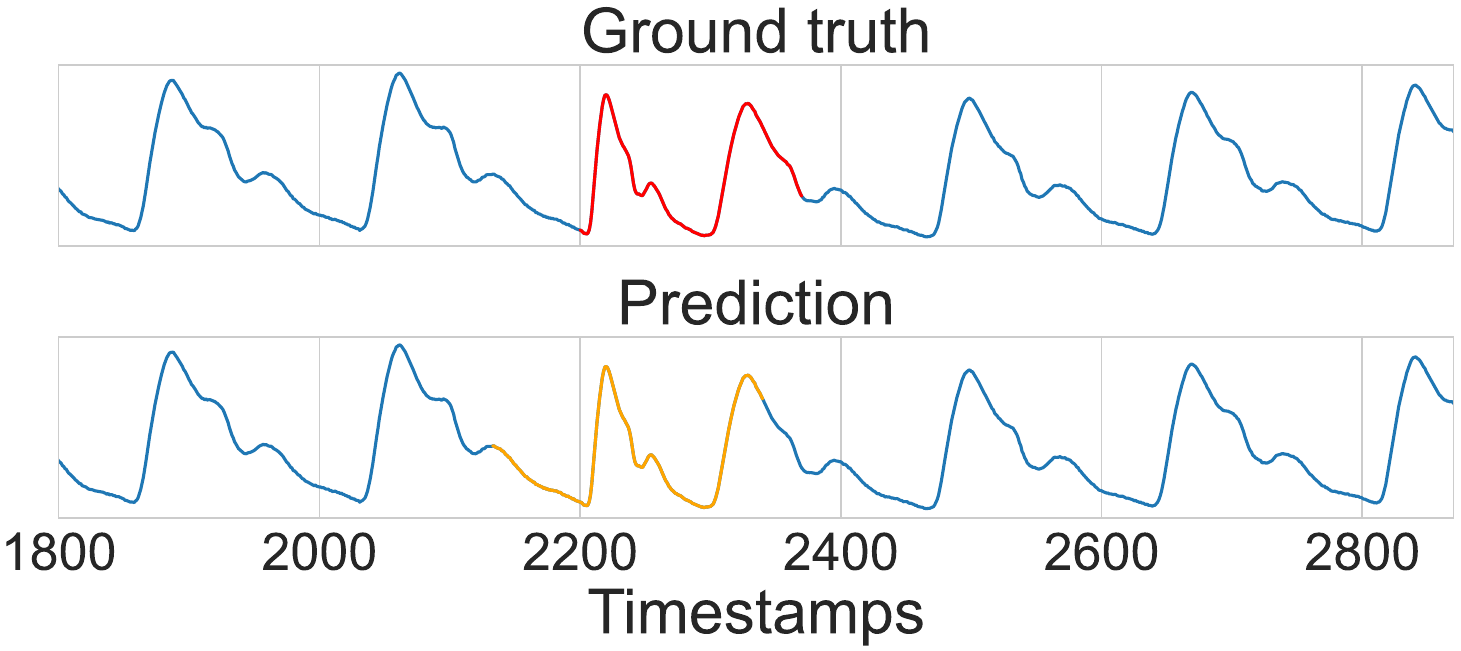}  
    \caption{UCR 141: Seasonal anomaly}
\end{subfigure}
\begin{subfigure}{.325\textwidth}
    \centering
    \includegraphics[width=\linewidth]{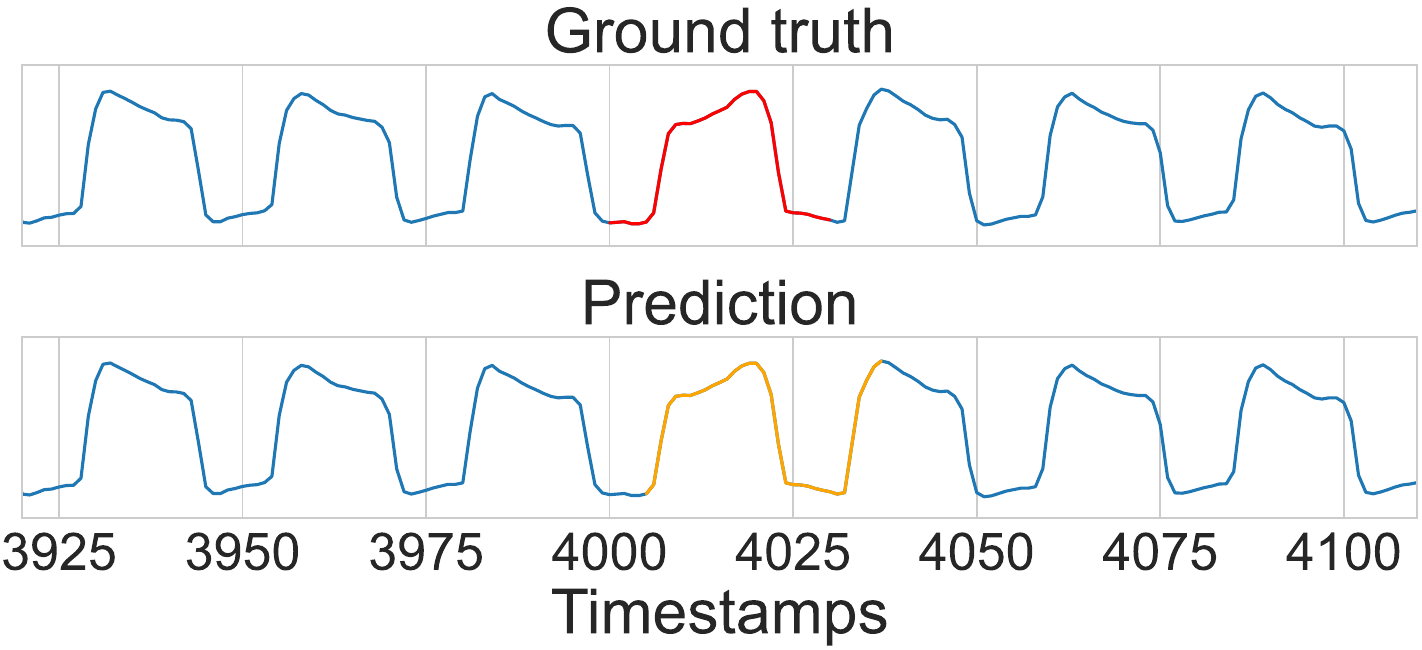}  
    \caption{UCR 173: Trend anomaly}
\end{subfigure}
\begin{subfigure}{.325\textwidth}
    \centering
    \includegraphics[width=\linewidth]{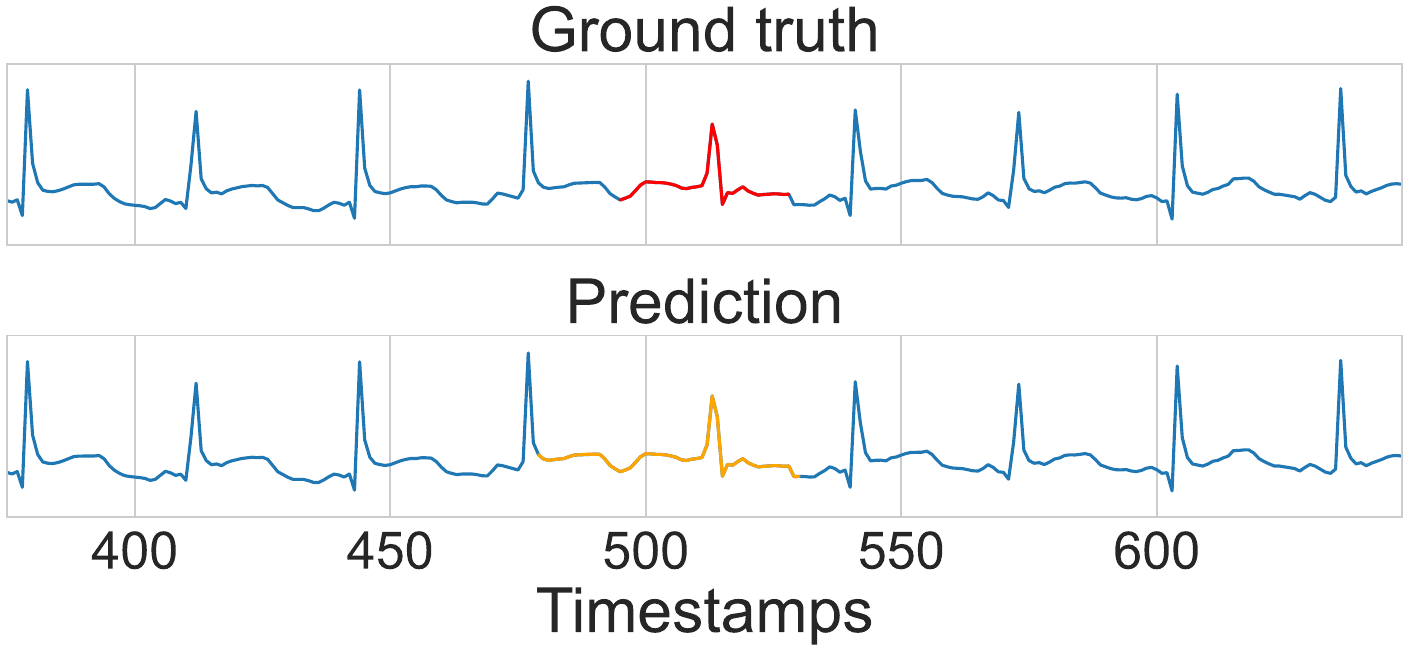}  
    \caption{UCR 203: Level shifts anomaly}
\end{subfigure}
\begin{subfigure}{.325\textwidth}
    \centering
    \includegraphics[width=\linewidth]{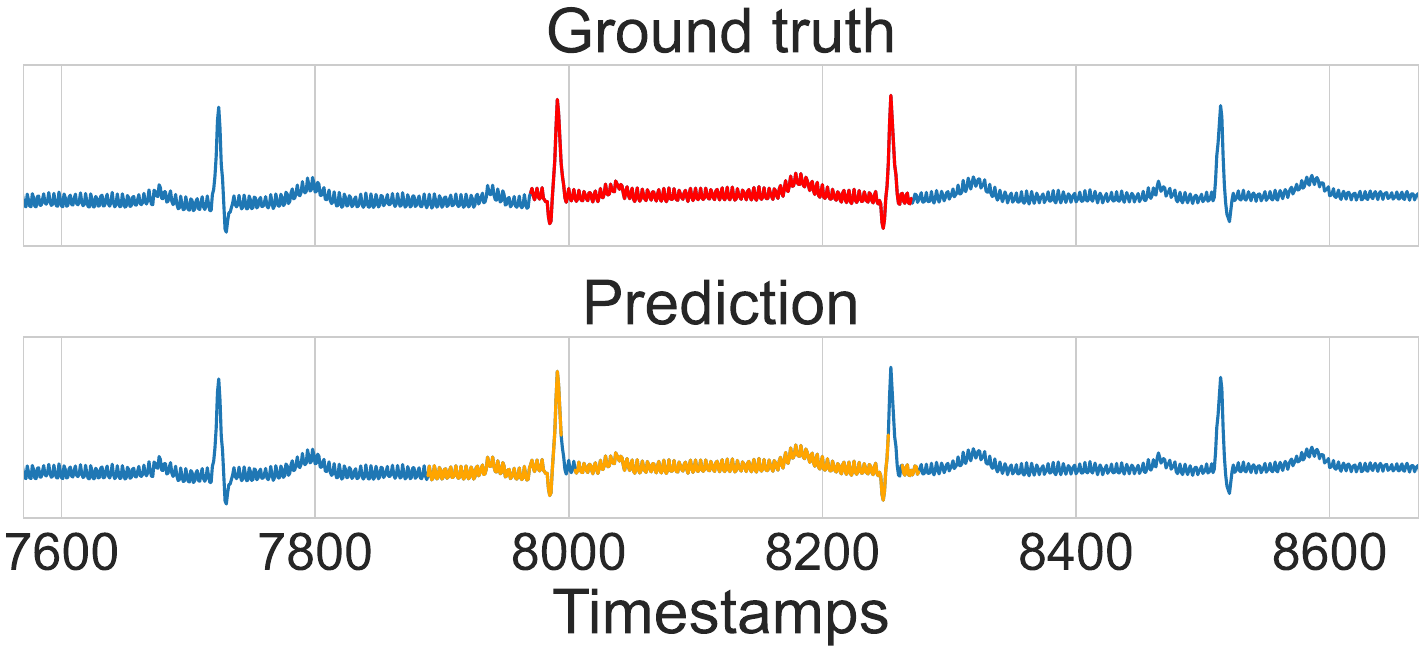}  
    \caption{UCR 209: Contextual anomaly}
\end{subfigure}
\caption{Various anomaly types detected by TriAD.}
\label{fig:anomaly_type}
\end{figure*}

In Fig.~\ref{fig:anomaly_type}, we showcase the proficiency of TriAD in detecting a diverse range of anomalies, irrespective of their length and shape. We highlight six distinct anomalies from the test set to evident this versatility: the noise-like anomaly, characterized by unexpected fluctuations; duration anomaly, denoting unexpected extensions of stable behavior; seasonal anomaly, where there's an abrupt doubling of inherent seasonality; trend anomaly that reveals an unanticipated rise; level shifts anomaly with occurrences of lasting shift characterized by a sudden jump or drop; and the contextual anomaly where normal sequences are distorted. These results clearly emphasize TriAD's capability in spotting anomalies ranging from lengths of 20 to 200, and demonstrate particular robustness against more subtle anomalies like duration, level shifts, and contextual anomalies. In contrast, as depicted in Fig.~\ref{fig:mtgflow}, the results from the primary competitor MTGFlow reveal a tendency to misclassify normal patterns as anomalies, which is attributed to its limitation in capturing distinctive features within subtle changes in the data. Hence, the diversity of detected anomalies further emphasizes TriAD's capability in distinguishing variations across temporal, frequency, and residual domains.

\subsection{Case when discord discovery fails}
\begin{figure}[hbt!]
  \centering
  \captionsetup{justification=centering}
  \includegraphics[width= \linewidth]{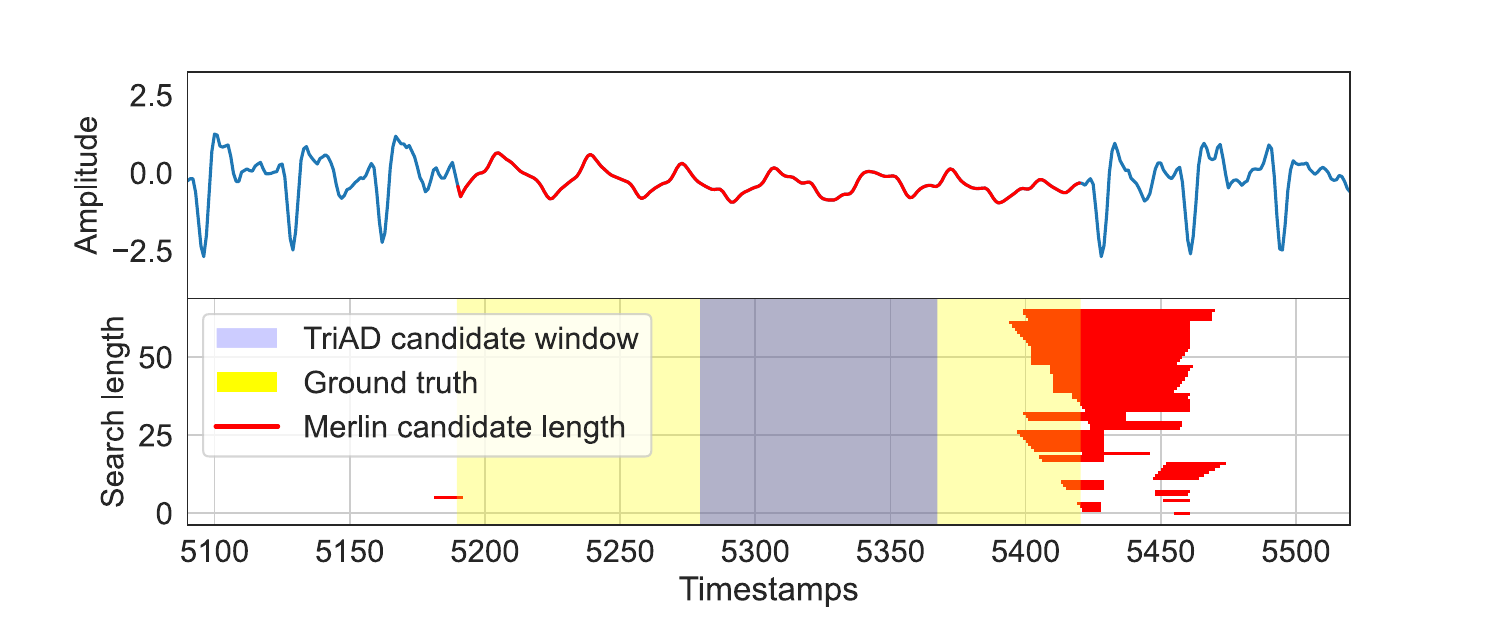}
  \caption{Example from UCR ``150": case when anomalous segment dominates the search window.}
  \label{fig:case_150}
\end{figure}

As evidenced in Fig.~\ref{fig:case_150}, while TriAD shows its superiority in discovering the most suspect window, the subsequent discord discovery might fail if the search window holds more anomalous data than normal data, especially when the anomalous events span a wide length. In such scenarios, discord discovery algorithms like MERLIN might misidentify regular patterns as anomalies, overlooking genuine anomalies. To rectify this, our proposed framework implements an exception: assigning all data points within the window as positives. The underlying reasoning is that if TriAD identifies a window as potentially containing anomalies and none are found, it's probable that regular patterns outside the window were wrongly flagged as anomalies.

\section{Related works}
\textbf{Supervised and unsupervised learning in time series anomaly detection}.
Various machine learning methodologies can detect anomalies using both supervised and unsupervised strategies. While supervised approaches deploy generative and discriminative models to distinguish between regular and abnormal patterns, the limited availability and inherent imbalance of labels often hinder their effective deployment \cite{schmidl2022anomaly, zhang2022aggregation}. Given these challenges, the focus has shifted to unsupervised techniques, which predict or reconstruct time-series data based on historical patterns and determine anomalies based on the divergence between its prediction and the actual data. Notably, CNNs and RNNs have gained distinction in unsupervised anomaly detection, often paired with autoencoders \cite{2018.2886457, TSMC.2020.2968516}. In contrast to RNN-centric architectures, attention mechanisms have demonstrated resilience in modeling sequential data, capitalizing on parallel processing capabilities \cite{8594829}. Inspired by the proficiency of self-attention in tracking extended temporal patterns, researchers also have introduced attention-centric models for anomaly detection, that leverage multi-modal features derived from extensive time sequences \cite{3514067, ding2023mst}. Additionally, there is a growing interest in real-time anomaly detection in Industrial Internet of Things (IIoT) systems. This approach addresses the challenges associated with constrained memory budgets \cite{10093068} and privacy concerns \cite{wang2021fast} related to the deployment of deep learning models.

\textbf{Self-supervised learning in time series}.
Addressing the challenges posed by the lack of labels in time series anomaly detection, self-supervised learning has made notable strides in various time series applications. A popular approach involves synthesizing anomalies and training models to differentiate them from regular instances \cite{xu2022calibrated,9705079, ijcai2022p394, jin2023trafformer}. For instance, NCAD \cite{ijcai2022p394} refines the distinction between normal and anomalous samples by incorporating single point outliers during its training phase. Beyond this, several studies have explored the potential of view generation, crafting multiple contrasting perspectives through multi-scale augmentations \cite{dcdetector23, 10001758}. For example, MTFCC \cite{10001758} novelly creates views drawing on multi-scale properties. By sampling time series across varied scales, it assumes that views derived from an identical sample should share similar representations, leading to the introduction of the multi-timescale feature consistency constraint during encoder training. However, the challenges of choosing the best augmentation strategies that preserve temporal dependencies and defining the respective positive and negative pairs still require further exploration \cite{zhang2023self, li2018neural}.

\section{Conclusion}
To address the two main challenges in time series anomaly detection - the limited availability of labels and the variations of anomalies - we introduce a label-independent framework TriAD using contrastive learning combined with carefully designed data augmentations. Furthermore, we tackle longstanding issues in benchmarking, offering a thorough, experiment-based comparison of the deep learning models using rigorous evaluations. Our experimental results reveal a remarkable threefold increase in optimized point-wise F1 scores over recent leading models. This breakthrough emphasizes the potential and reliability of deep learning models in serving as effective tools for future time series anomaly detection tasks, which adds to the ongoing debate about the effectiveness of deep learning techniques in this area.

\section*{Acknowledgment}
We gratefully thank Dr. Radislav Vaisman and Dr. Thomas Taimre (the University of Queensland). Their invaluable input and guidance are crucial to the completion of this paper. During moments of challenges, their support and encouragement provided the motivation we needed to persevere.

\bibliographystyle{IEEEtran}
\bibliography{citation}
\end{document}